\pdfoutput=1

\documentclass[11pt]{article}

\usepackage[preprint]{acl}

\usepackage{times}
\usepackage{latexsym}

\usepackage[T1]{fontenc}

\usepackage[utf8]{inputenc}

\usepackage{microtype}

\usepackage{inconsolata}

\usepackage{graphicx}
\usepackage{mdframed}
\usepackage{amsmath}
\usepackage{multirow}
\usepackage{overpic}
\usepackage{bbding}
\usepackage{booktabs} 
\usepackage{subcaption}
\usepackage{float}
\definecolor{myblue}{RGB}{56,94,124}
\definecolor{myred}{RGB}{176,35,24}
\definecolor{mygreen}{RGB}{76,123,49}
\definecolor{textred}{RGB}{176,35,24}

\title{Mitigating Hallucinations in Multimodal Spatial Relations through Constraint-Aware Prompting}

\author{Jiarui Wu \\
  University of Rochester \\
  \texttt{jwu114@u.rochester.edu} \\\And
  Zhuo Liu \\
  University of Rochester \\
  \texttt{zhuo.liu@rochester.edu} \\\And
  Hangfeng He \\
  University of Rochester \\
  \texttt{hangfeng.he@rochester.edu} \\}

\begin{document}
\maketitle
\begin{abstract}

Spatial relation hallucinations pose a persistent challenge in large vision-language models (LVLMs), leading to generate incorrect predictions about object positions and spatial configurations within an image. To address this issue, we propose a constraint-aware prompting framework designed to reduce spatial relation hallucinations. Specifically, we introduce two types of constraints: (1) bidirectional constraint, which ensures consistency in pairwise object relations, and (2) transitivity constraint, which enforces relational dependence across multiple objects. By incorporating these constraints, LVLMs can produce more spatially coherent and consistent outputs. We evaluate our method on three widely-used spatial relation datasets, demonstrating performance improvements over existing approaches.\footnote{Our code is available at \href{https://github.com/jwu114/CAP}{https://github.com/jwu114/CAP}.} Additionally, a systematic analysis of various bidirectional relation analysis choices and transitivity reference selections highlights greater possibilities of our methods in incorporating constraints to mitigate spatial relation hallucinations.


\end{abstract}

\begin{figure}[t]
    \centering
    \fbox{%
    \begin{overpic}[width=0.94\linewidth]{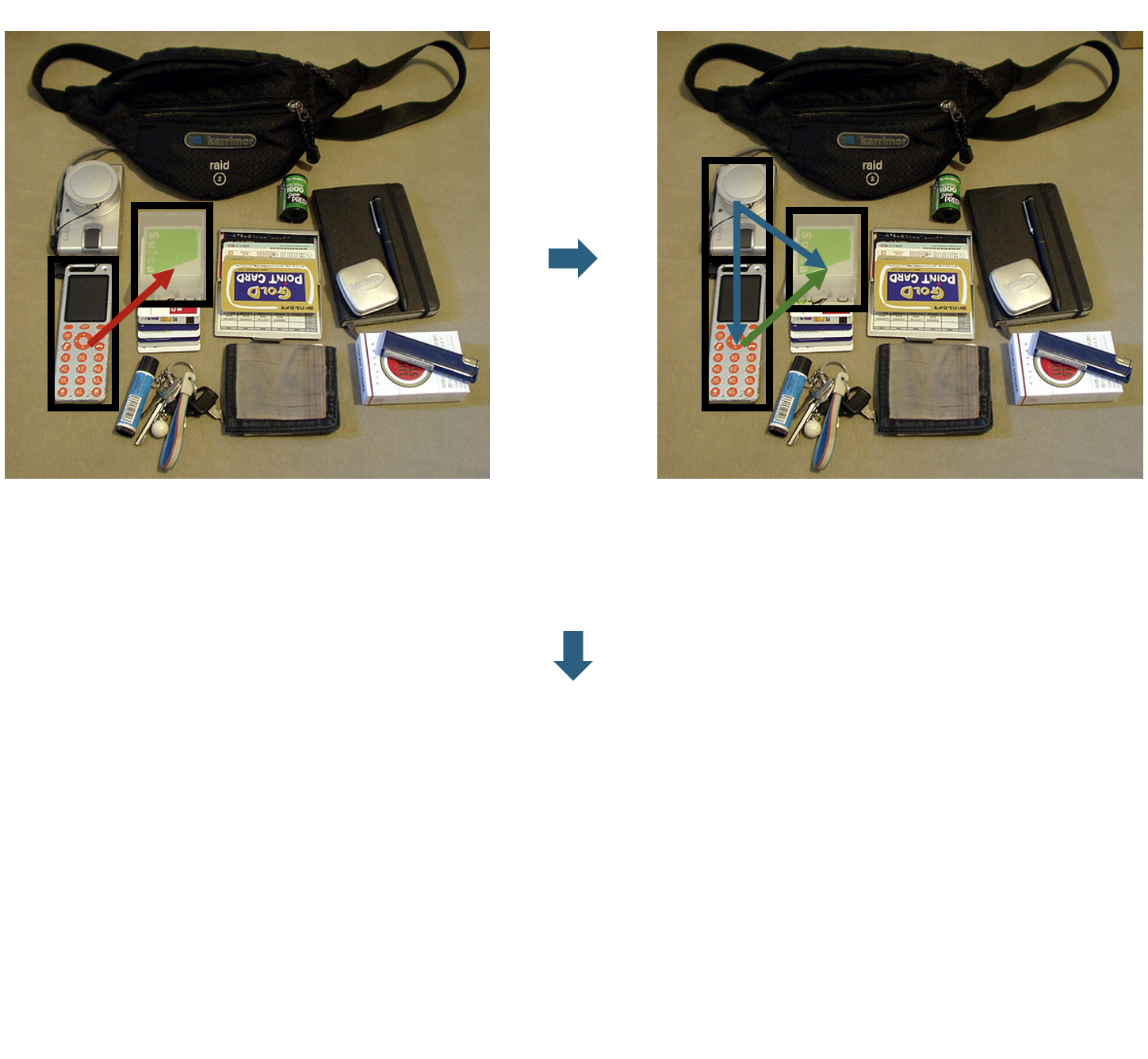}
        \put(6.7,65.2){\bfseries{
        \begin{minipage}{0.91\linewidth}   
        \linespread{0.8}\fontsize{5}{12}\raggedright\selectfont\sloppy\textcolor{myred}{
            \rotatebox{41}{Left}
        }
        \end{minipage}}}
        \put(63.3,65.2){\bfseries{
        \begin{minipage}{0.91\linewidth}   
        \linespread{0.8}\fontsize{5}{12}\raggedright\selectfont\sloppy\textcolor{mygreen}{
            \rotatebox{41}{Right}
        }
        \end{minipage}}}
        \put(63.7,72){\bfseries{
        \begin{minipage}{0.91\linewidth}   
        \linespread{0.8}\fontsize{5}{12}\raggedright\selectfont\sloppy\textcolor{myblue}{
            \rotatebox{-36}{Right}
        }
        \end{minipage}}}
        \put(60.5,68){\bfseries{
        \begin{minipage}{0.91\linewidth}   
        \linespread{0.8}\fontsize{5}{12}\raggedright\selectfont\sloppy\textcolor{myblue}{
            \rotatebox{90}{Below}
        }
        \end{minipage}}}
        \put(-1,22){\bfseries{
        \begin{minipage}{0.91\linewidth}   \linespread{0.8}\fontsize{7}{12}\raggedright\selectfont\sloppy 
            Q: Is the transparent card holder on the right of the cell phone? 
            \\A: \textcolor{myred}{No, the transparent card holder is on the left of the cell phone. \XSolidBrush}\vspace{12pt}
            Q: \textcolor{myblue}{What is the spatial relation between the transparent card holder and the camera and between the cell phone and the camera?} Based on the answer to the previous question, is the transparent card holder on the right of the cell phone?
            \\A: \textcolor{myblue}{The transparent card holder is to the right of the camera. The cell phone is below the camera.} Based on this,  \textcolor{mygreen}{the transparent card holder is to the right of the cell phone.\qquad\qquad\qquad\qquad\qquad\;\;\; \CheckmarkBold}
        \end{minipage}}}
    \end{overpic}
    }
    \caption{Comparison between the vanilla prompt and the prompt incorporating constraint awareness (transitivity constraint). Constraint-aware content is highlighted in blue, incorrect content in red, and correct content in green. In the right image, the relations highlighted in blue corrects the incorrect relation highlighted in red.}
    \label{fig:demo}
\end{figure}

\section{Introduction}

In recent years, large vision-language models (LVLMs) have been widely adopted for tasks such as image captioning and visual question answering (VQA). While these models have demonstrated remarkable capabilities, hallucination remains a persistent challenge in multimodal systems. Even state-of-the-art models occasionally generate hallucinated responses~\citep{chang2024survey}. In this study, we focus on mitigating the hallucination in multimodal spatial relations, a challenging task that requires the cognition and reasoning ability of LVLMs about objects in the image. 

Existing research has explored various methods to enhance the performance of LVLMs in spatial relations.~\citet{zhao2023enhancing} utilized a small pretrained model to provide spatial information in guiding LVLMs.~\citet{rajabi2023towards} combined an encoder-decoder model with a trained predictor to localize objects and predict spatial relations.~\citet{chen2024spatialvlm} proposed SpatialVLM, trained on a spatial VQA dataset generated using their data generation framework. Additionally,~\citet{meng2024know} introduced ZeroVLM, which leverages a 3D reconstruction model to obtain multi-view images for improved spatial reasoning. The existing proposals mainly focus on training powerful models. While these approaches are effective in enhancing LVLMs' visual spatial relation understanding, they involve high training costs and rely heavily on high-quality training data.

Prompt enhancement is a training-free approach that has been shown to effectively mitigate hallucinations in large language models (LLMs). Numerous prompting techniques (e.g.,~\citealp{NEURIPS2022_9d560961,hu2023chain,kong-etal-2024-better,zheng2024take}) have been developed to improve response quality in reasoning tasks. However, despite their success, few methods have achieved significant improvements in multimodal spatial relation tasks~\citep{sahoo2024systematic, vatsal2024survey}.

In response to these limitations, we propose constraint-aware methods that effectively reduce spatial relation hallucinations of LVLMs. These methods are inspired by the principle that, in tasks involving structured variables, once a variable's value is determined, related variables become constrained~\citep{ning-etal-2019-partial}. Specifically, in spatial relation reasoning, establishing the spatial relationship between two objects naturally constrains the potential relationships among other objects in the scene~\citep{choi2018structured}. For instance, Figure~\ref{fig:demo} demonstrates a scenario~\citep{yang2019spatialsense} where the model initially misinterprets the spatial relation between the transparent card holder and the cell phone. By identifying the relations among the camera, the card holder, and the cell phone, the initially incorrect relation is constrained and corrected. 

We propose two constraints: bidirectional constraint and transitivity constraint. Bidirectional constraints ensure that the spatial relations between two objects remain consistent when viewed from either direction. Transitivity constraints, introduce a third object as a reference to maintain logical coherence across multiple spatial relations, reducing the likelihood of conflicting interpretations. By combining these constraints, we establish a more robust approach for visual spatial relation.


We compare our methods against baseline methods using three widely used spatial relation datasets. The results show that all our methods have significantly improved performance, with the combined method outperforming the other two constraint methods. These findings demonstrate the effectiveness of our approach in mitigating multimodal spatial relation hallucinations. Furthermore, we analyze the performance across different method variants, highlighting their effectiveness and the variability in performance across datasets.

\section{Constraint-Aware Prompting}
This section provides a description of our proposed methods. The demonstrated methods are designed for spatial relation binary VQA, where the input is typically an image-question pair. The underlying constraint-aware approach has the potential to be applied to a broader range of spatial relation tasks.

\begin{figure}[t]
    \centering
    \begin{overpic}[width=1\linewidth]{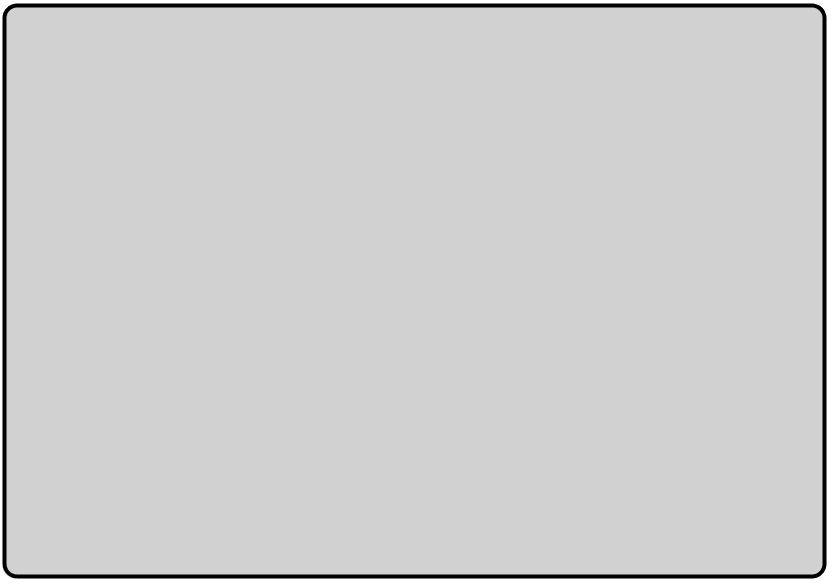}
        \put(4,33.5){\textcolor{black}{
        \begin{minipage}{0.91\linewidth}
            \fontsize{8}{12}\selectfont
            \textbf{\#\# Instructions \#\#}
            \\1. Repeat the question + \{extract and label objects\}
            \\2. \{spatial relation analysis\}
            \\3. \textcolor{blue}{Think step by step} + use "yes" or "no" to answer the question
            \\\textbf{\#\# Please output in the following format \#\#}
            \\...
            \\\textcolor{blue}{Horizontal relation} between $O_1$ and $O_2$: $O_1$ is <relation> $O_2$
            \\\textcolor{blue}{Vertical relation} between $O_1$ and $O_2$: $O_1$ is <relation> $O_2$
            \\\textcolor{blue}{Depth relation} between $O_1$ and $O_2$: $O_1$ is <relation> $O_2$
            \\...
            \\\textbf{\#\# Question \#\#}
            \\Is there <Object> <Relation> <Object> in the image?
        \end{minipage}}}
    \end{overpic}
    \caption{Template prompt skeleton. Prompting techniques are highlighted in blue. The phrase inside \{\} is the summary of omitted details, and $O_1$ and $O_2$ represent the label of objects.} 
    \label{fig:prompt8}
\end{figure}

\begin{figure}[t]
    \centering
    \fbox{%
    \begin{overpic}[width=0.94\linewidth]{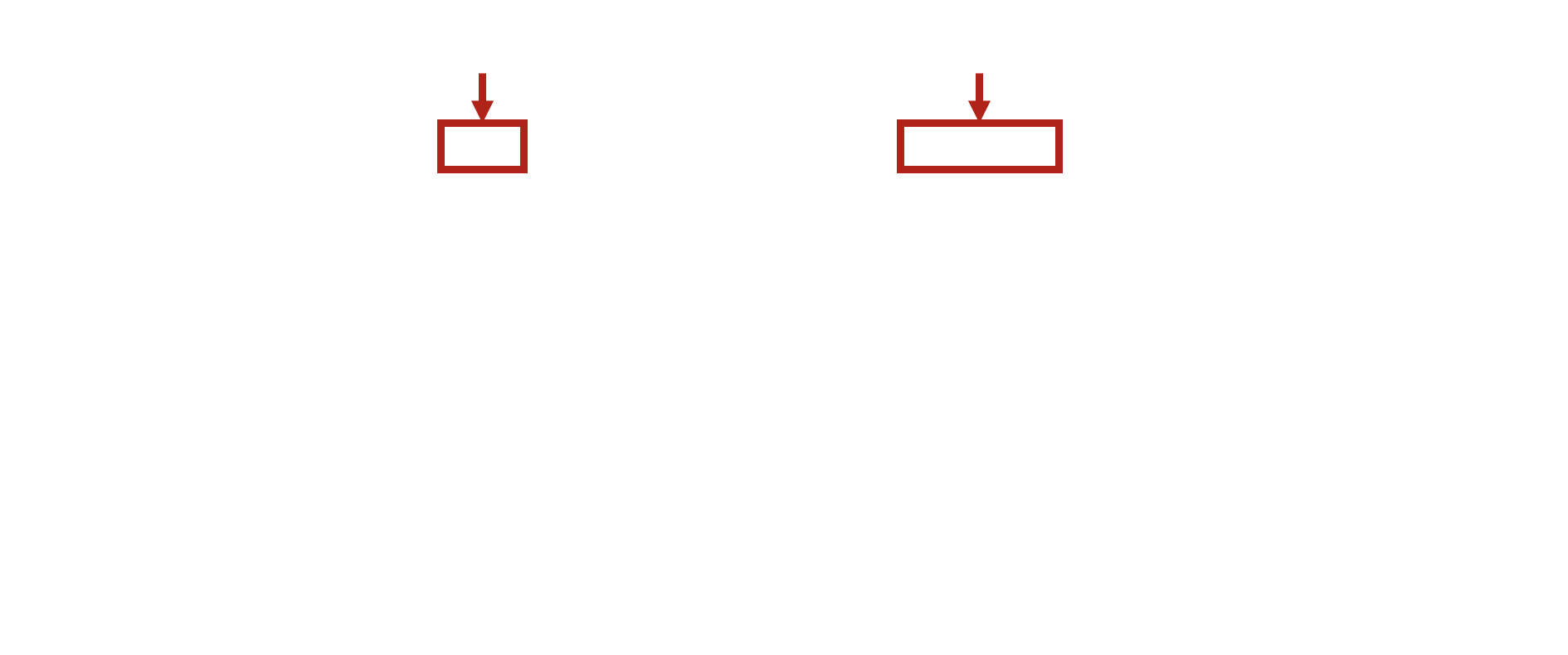}
        \put(28.5,38){\bfseries{
        \begin{minipage}{0.91\linewidth}   
        \linespread{0.8}\fontsize{7}{12}\raggedright\selectfont\sloppy\textcolor{myred}{A}
        \end{minipage}}}
        \put(60,38){\bfseries{
        \begin{minipage}{0.91\linewidth}   
        \linespread{0.8}\fontsize{7}{12}\raggedright\selectfont\sloppy\textcolor{myred}{B}
        \end{minipage}}}
        \put(-2.2,15.8){
        \begin{minipage}{0.96\linewidth}   \linespread{0.8}\fontsize{7}{12}\raggedright\selectfont\sloppy 
            \textbf{Question: Is there a cat on the right of a rabbit in the image?} \\\vspace{6pt}
            \textbf{AB Relation:} What is the relation between A (cat) and B (rabbit)?\\
            \textbf{BA Relation:} What is the relation between B (rabbit) and A (cat)?\\
            \textbf{AB+BA Relation:} What is the relation between A (cat) and B (rabbit) and between B (rabbit) and A (cat)?\\
            \textbf{BA+AB Relation:} What is the relation between B (rabbit) and A (cat) and between A (cat) and B (rabbit)?\\
        \end{minipage}}
    \end{overpic}
    }
    \caption{Example shows how candidate objects in the question are labeled and the corresponding spatial relations in the AB, BA, AB+BA, and BA+AB orders. "Cat" is labeled as "A" because it appears earlier than "rabbit" in the question.}
    \label{fig:ab_example}
\end{figure}

\begin{table*}[t]
\centering
\scalebox{0.79}{
    \begin{tabular}{l c c c c c c c c}
    \toprule
     & \multicolumn{2}{c}{\textbf{ARO}} & \multicolumn{2}{c}{\textbf{GQA}} & \multicolumn{2}{c}{\textbf{MMRel}} & \multicolumn{2}{c}{\textbf{Average}} \\
    \textbf{Methods} & \textbf{Acc} & \textbf{F1} & \textbf{Acc} & \textbf{F1} & \textbf{Acc} & \textbf{F1} & \textbf{Acc} & \textbf{F1} \\
    \midrule
    Baseline (vanilla) & 65.10 & 70.75 & 63.03 & 67.73 & 70.96 & 75.12 & 66.37 & 71.20\\ 
    Baseline (CoT+structure) & 69.60 & 74.51 & 63.47 & 68.03 & 80.00 & 81.41 & 71.02 & 74.65\\ 
    Bidirectional (ours) & 75.33\ \raisebox{0.5ex}{\textasteriskcentered{}\textasteriskcentered{}\textasteriskcentered{}} & 79.03\ \raisebox{0.5ex}{\textasteriskcentered{}\textasteriskcentered{}\textasteriskcentered{}} & 69.47\ \raisebox{0.5ex}{\textasteriskcentered{}\textasteriskcentered{}\textasteriskcentered{}} & 73.86\ \raisebox{0.5ex}{\textasteriskcentered{}\textasteriskcentered{}\textasteriskcentered{}} & 88.50\ \raisebox{0.5ex}{\textasteriskcentered{}\textasteriskcentered{}\textasteriskcentered{}} & 89.12\ \raisebox{0.5ex}{\textasteriskcentered{}\textasteriskcentered{}\textasteriskcentered{}} & 77.77\ \raisebox{0.5ex}{\textasteriskcentered{}\textasteriskcentered{}\textasteriskcentered{}} & 80.67\ \raisebox{0.5ex}{\textasteriskcentered{}\textasteriskcentered{}\textasteriskcentered{}}\\ 
    Transitivity (ours) & 73.90\ \raisebox{0.5ex}{\textasteriskcentered{}\textasteriskcentered{}\textasteriskcentered{}} & 75.89\ \raisebox{0.5ex}{\textasteriskcentered{}\textasteriskcentered{}\textasteriskcentered{}} & 68.93\ \raisebox{0.5ex}{\textasteriskcentered{}\textasteriskcentered{}\textasteriskcentered{}} & 71.04\ \raisebox{0.5ex}{\textasteriskcentered{}\textasteriskcentered{}\textasteriskcentered{}} & 84.03\ \raisebox{0.5ex}{\textasteriskcentered{}\textasteriskcentered{}\textasteriskcentered{}} & 83.19\ \raisebox{0.5ex}{\textasteriskcentered{}\textasteriskcentered{}\textasteriskcentered{}} & 75.62\ \raisebox{0.5ex}{\textasteriskcentered{}\textasteriskcentered{}\textasteriskcentered{}} & 76.71\ \raisebox{0.5ex}{\textasteriskcentered{}\textasteriskcentered{}\textasteriskcentered{}}\\ 
    Combined (ours) & \textbf{76.67\ }\raisebox{0.5ex}{\textasteriskcentered{}\textasteriskcentered{}\textasteriskcentered{}} & \textbf{79.57\ }\raisebox{0.5ex}{\textasteriskcentered{}\textasteriskcentered{}\textasteriskcentered{}} & \textbf{70.77\ }\raisebox{0.5ex}{\textasteriskcentered{}\textasteriskcentered{}\textasteriskcentered{}} & \textbf{75.08\ }\raisebox{0.5ex}{\textasteriskcentered{}\textasteriskcentered{}\textasteriskcentered{}} & \textbf{92.70\ }\raisebox{0.5ex}{\textasteriskcentered{}\textasteriskcentered{}\textasteriskcentered{}} & \textbf{92.91\ }\raisebox{0.5ex}{\textasteriskcentered{}\textasteriskcentered{}\textasteriskcentered{}} & \textbf{80.05\ }\raisebox{0.5ex}{\textasteriskcentered{}\textasteriskcentered{}\textasteriskcentered{}} & \textbf{82.52\ }\raisebox{0.5ex}{\textasteriskcentered{}\textasteriskcentered{}\textasteriskcentered{}}\\ 
    
    \bottomrule
    \end{tabular}
}\caption{5-trial average results of our methods on three datasets using GPT-4o. The "average" column represents the overall performance across the datasets.  \raisebox{0.5ex}{\textasteriskcentered{}\textasteriskcentered{}\textasteriskcentered{}} indicates that the p-value of the one-sided t-test is less than 0.05 (comparing our methods with others and comparing the combined constraint with the other two constraints).}
\label{tab:main_results}
\end{table*}

\paragraph{Skeleton}
\label{sec:skeleton}
\noindent As illustrated in Figure~\ref{fig:prompt8}, our methods follow the structure: \textit{Instructions + Output Format + Question}. Instead of relying on a few-shot prompt, we use a zero-shot prompt with step-by-step instructions to reduce costs and specify an output format to facilitate validation and evaluation.

To minimize hallucination in intermediate steps and enhance LVLM reasoning, we incorporate various techniques. We leverage zero-shot chain-of-thought (CoT) prompting~\citep{NEURIPS2022_9d560961} to enable LVLMs to reason effectively based on detected spatial relations. In the output format, we adopt a reasoning structure~\citep{zhou2024self}, explicitly instructing LVLMs to analyze horizontal, vertical, and depth relations between objects. This approach ensures that models generate comprehensive spatial relations and engage in thorough reasoning. A detailed analysis of these techniques can be found in Appendix~\ref{sec:appendix1}.

Our methods also instruct LVLMs to label the first and second objects mentioned in the candidate VQA question as A and B, respectively. These symbols are then used to guide the models in the subsequent spatial relation analysis. An example is illustrated in Figure~\ref{fig:ab_example}.


\paragraph{Bidirectional Constraint}
\noindent In the method, we prompt LVLMs to generate spatial relations in the \textit{BA + AB} order. This approach ensures that LVLMs first detect the converse spatial relation (BA) and automatically refer to it when generating the direct relation (AB). The bidirectional constraint between the converse relation and the direct relation can help models mitigate hallucinations. The example response can be found in Figure~\ref{fig:example1} and the detailed template prompt can be found in Figure~\ref{fig:prompt1} in Appendix.

\paragraph{Transitivity Constraint}
\noindent This method leverages the transitivity constraint among objects to mitigate spatial relation hallucinations. Besides Object A and B, we prompt LVLMs to randomly select a reference object, denoted as Object C. The model is then instructed to generate the spatial relations in the \textit{AC + BC} order, which serve as reference relations to transitively constrain the potentially hallucinated AB relation. The example response can be found in Figure~\ref{fig:example2} and the detailed template prompt can be found in Figure~\ref{fig:prompt2} in Appendix.


\section{Experiments}
\subsection{Experimental Settings}
\label{sec:settings}
\paragraph{Datasets} 
\noindent We utilize three datasets containing spatial relation data to evaluate our proposed methods. ARO~\citep{yuksekgonul2023and} consists of 50K real-world image-caption pairs, with data sourced from Visual Genome~\citep{krishna2017visual}, MSCOCO~\citep{lin2014microsoft}, and Flickr30k~\citep{young2014image}. GQA~\citep{hudson2019gqa} includes 113K images and 22M diverse visual questions based on the Visual Genome scene graph. MMRel~\citep{nie2024mmrel} contains 15K image-question pairs addressing GPT-4V-generated annotations, utilizing real images from Visual Genome and synthetic images from SDXL~\citep{podell2024sdxl} and DALL-E~\citep{betker2023improving}. We randomly sampled 600 and 200 balanced pairs of real images and spatial relation binary VQA questions from each dataset for the test and validation splits. Data preprocessing details are in Appendix~\ref{sec:appendix4}.

\paragraph{Methods} 
With the hypothesis that the combination of bidirectional and transitivity constraints can yield improved performance, we introduce the combined constraint. It integrates two constraints and performs the relation analysis in the \textit{AC + BC + BA + AB} order. In the experiment, we use the above three constraint-aware methods against two baseline methods. The first baseline uses vanilla prompts, directly asking LVLMs to answer  questions with either "yes" or "no." The other baseline is based on the vanilla prompts but leverages prompting techniques, such as CoT and structured reasoning output (CoT+structure), which are also incorporated into our methods. The detailed prompts can be found in Appendix~\ref{sec:appendix7}. For each method and dataset, we conduct five trials, calculate the average results, and perform a one-sided t-test to further ensure the reliability of the findings.

\paragraph{Model Settings}
\noindent We use GPT-4o\footnote{GPT-4o-2024-05-13}~\citep{radford2018improving} as the LVLM in all experiments. The temperature and top-p are both set to a small number $1 \times 10^{-15}$, and a fixed seed is used to get more deterministic responses.

\begin{figure*}[t]
    \centering
    \includegraphics[width=\textwidth]{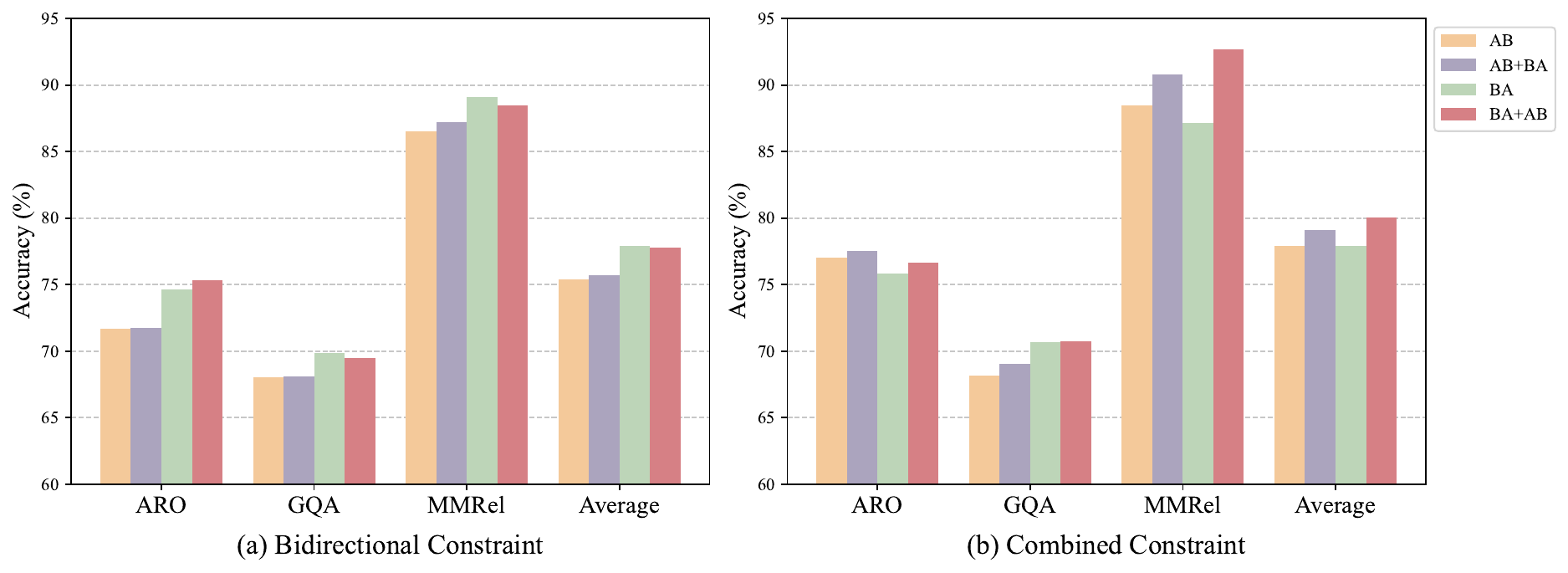}
    \caption{The accuracy comparison of different relation analysis choices in bidirectional and combined constraints is shown. \textit{BA + AB} is the method utilized in our proposed approach. \textit{BA} and \textit{AB + BA} are the variants of our method: \textit{BA} refers to analyzing only the converse relation, while \textit{AB + BA} analyzes the direct relation first, followed by the converse relation. \textit{AB}, which only analyzes the direct relation, is not considered a bidirectional constraint, as the converse relation is not examined. For the diagram of F1 score and detailed data, refer to Appendix~\ref{sec:appendix2}.}
    \label{fig:relation}
\end{figure*}

\subsection{Results}
\noindent As shown in Table~\ref{tab:main_results}, all of our proposed methods significantly outperform two baselines in both accuracy and F1 score. This highlights the effectiveness of our constraint-aware methods. Moreover, the combined constraint achieves the highest accuracy and F1 score across the three datasets, including an impressive 92.7\% accuracy and 92.91\% F1 score on the MMRel dataset. This demonstrates that the features of bidirectional and transitivity constraints can be combined to enhance performance.

To prove that our methods are generalized to other models, we also test them using different LVLMs. The results can be found in Appendix~\ref{sec:additional_appendix}.

\section{Analysis}
\subsection{Bidirectional Relation Analysis}
\label{sec:analysis1}
When utilizing the bidirectional constraint, the key factor is ensuring that LVLMs analyze the converse relation (\textit{BA}). In our experiments, we prompted GPT to analyze spatial relations in the \textit{BA + AB} order. However, alternative approaches exist, such as analyzing in the \textit{AB + BA} order or focusing solely on \textit{BA}.

We compare the performance of these variants under bidirectional and combined constraints. As shown in Figure~\ref{fig:relation}, methods analyzing the \textit{BA} relation demonstrated higher average accuracy compared to those using only the \textit{AB} relation, reaffirming the effectiveness of bidirectional constraints. However, no single method consistently outperformed the others across datasets, with results varying between them. Additionally, we observed that within the group employing bidirectional constraints, the \textit{AB} and \textit{AB + BA} performed similarly, as did the \textit{BA} and \textit{BA + AB}. We hypothesize that when the first relation is hallucinated, the subsequent relation is likely to be affected as well. This aligns with the findings that when LLMs reach a hallucinated answer, its subsequent explanations are also likely to be false~\citep{ye2022unreliability}.

\begin{table}[t]
\centering
\scalebox{0.56}{
    \begin{tabular}{l c c c c c c c c}
    \toprule
     & \multicolumn{2}{c}{\textbf{ARO}} & \multicolumn{2}{c}{\textbf{GQA}} & \multicolumn{2}{c}{\textbf{MMRel}} & \multicolumn{2}{c}{\textbf{Average}}\\
    \textbf{Attributes} & \textbf{Acc} & \textbf{F1} & \textbf{Acc} & \textbf{F1} & \textbf{Acc} & \textbf{F1} & \textbf{Acc} & \textbf{F1}\\
    \midrule
    The largest & \textbf{75.00} & 76.92 & 65.00 & 67.49 & 83.83 & \textbf{83.25} & 74.61 & 75.89 \\ 
    The smallest & 73.33 & 75.83 & 66.00 & 67.92 & 84.00 & 82.98 & 74.44 & 75.58 \\ 
    The most top & 72.17 & 73.62 & 65.50 & 67.91 & 81.67 & 80.84 & 73.11 & 74.12 \\ 
    The central & \textbf{75.00} & 76.78 & \textbf{69.17} & 70.77 & 83.00 & 81.72 & \textbf{75.72} & 76.42 \\ 
    The most obvious & \textbf{75.00} & \textbf{77.06} & 67.33 & 69.38 & 81.83 & 80.71 & 74.72 & 75.72 \\ 
    Random & 73.90 & 75.89 & 68.93 & \textbf{71.04} & \textbf{84.03} & 83.19 & 75.62 & \textbf{76.71} \\ 
    \bottomrule
    \end{tabular}
}\caption{The comparison of different reference object selection strategies in transitivity constraints.}
\label{tab:analysis1}
\end{table}

\begin{table}[t]
\centering
\scalebox{0.56}{
    \begin{tabular}{l c c c c c c c c}
    \toprule
     & \multicolumn{2}{c}{\textbf{ARO}} & \multicolumn{2}{c}{\textbf{GQA}} & \multicolumn{2}{c}{\textbf{MMRel}} & \multicolumn{2}{c}{\textbf{Average}}\\
    \textbf{Attributes} & \textbf{Acc} & \textbf{F1} & \textbf{Acc} & \textbf{F1} & \textbf{Acc} & \textbf{F1} & \textbf{Acc} & \textbf{F1}\\
    \midrule
    The largest & 77.00 & 79.53 & 70.50 & 75.04 & 91.67 & 91.88 & 79.72 & 82.15 \\ 
    The smallest & 75.83 & 79.02 & 69.67 & 74.37 & 91.17 & 91.41 & 78.89 & 81.60 \\ 
    The most top & 76.17 & 79.25 & 68.83 & 73.48 & 92.33 & 92.60 & 79.11 & 81.78 \\ 
    The central & \textbf{78.17} & \textbf{80.82} & 68.83 & 73.62 & 90.67 & 90.91 & 79.22 & 81.78\\ 
    The most obvious & 77.67 & 80.41 & 69.67 & 73.93 & 92.33 & 92.48 & 79.89 & 82.27 \\ 
    Random & 76.67 & 79.57 & \textbf{70.77} & \textbf{75.08} & \textbf{92.70} & \textbf{92.91} & \textbf{80.05} & \textbf{82.52}\\ 
    \bottomrule
    \end{tabular}
}\caption{The comparison of different reference object selection strategies in combined constraints.}
\label{tab:analysis2}
\end{table}

\subsection{Reference Selection}
\label{sec:analysis2}
The reference object in transitivity constraints plays a crucial role. Ideally, it should not introduce new hallucinations and must be tactically positioned to challenge the originally hallucinated relation. Thus, selecting a reliable reference object is essential.

In this analysis, we evaluate different reference selection strategies  using transitivity and combined constraints. This is done through prompting "select <attribute> object" as shown in Figures~\ref{fig:prompt6} and~\ref{fig:prompt7}. The attributes are manually defined as "the largest," "the smallest," "the most top," "the central," and "the most obvious," based on the assumption that they could affect the quality of reference objects.

As shown in Tables~\ref{tab:analysis1} and~\ref{tab:analysis2}, certain datasets demonstrate a preference for specific attributes. For instance, "the central" performs better in the ARO dataset, while random selection is better in the other two datasets. Although all strategies help mitigate spatial relation hallucinations, their performance varies depending on the dataset data distribution.

Generally, we found that the largest reference object performs slightly better than the smallest reference object, and an effective reference object often challenges the originally hallucinated relation. An important insight for selecting the third object is that a large object positioned between the candidate objects is more likely to challenge the hallucinated relation and mitigate the hallucination.

\section{Conclusion}
We have proposed two powerful constraint-aware methods, bidirectional and transitivity constraints, based on the inter-constraint relations among structured variables. These methods and their combinations and variants can be easily implemented to enhance the performance of LVLMs in visual spatial relations. We hope our proposed methods and experimental results can inspire further exploration in multimodal spatial relation tasks.

\section*{Limitations}
The proposed methods are inspired by constraints found in structured variables. This insight was examined and evaluated using binary VQA, and we did not evaluate our proposals with other visual tasks. Besides that, in our current evaluation, we primarily focus on regular spatial relationships, such as horizontal and depth relations. However, real-world data encompasses a broader range of spatial relations. They will be solid aspects for us to extend in future experiments.

For future research, we aim to develop a more deterministic automatic reference object selection mechanism to replace the current random selection used in the transitivity constraint. Appendix~\ref{sec:tradeoff_analysis} shows the tradeoff between accuracy and cost across the methods. Although our methods are highly cost-efficient, we can further reduce API costs while maintaining accuracy in the following research. Appendix~\ref{sec:appendix6} analyzes several failed cases of our current methods, offering insights that reveal potential directions for improving our methods in the future.

\bibliographystyle{acl_natbib}

\clearpage
\appendix

\section{Appendix}
\subsection{Data Preprocessing} 
\label{sec:appendix4}
We preprocess the datasets to align with our experimental objectives. To focus on more challenging data, GPT-4o-mini\footnote{GPT-4o-mini-2024-07-18} is used to filter the data, ensuring that all candidate images contain at least five objects. Non-spatial relations and vague spatial relations, such as "sitting on" and "nearby," are eliminated, leaving only those with clear spatial definitions in the images. Additionally, we format the annotations so that the questions follow the structure: \textit{Is there <Object> <Relation> <Object> in the image?} This makes image-question pair have uniform formats and allows us to focus on evaluating the feasibility of the proposed method in spatial relation tasks.

In details, the preprocessing of the ARO and GQA datasets is more complex than that of MMRel, as the annotations of ARO are primarily image captions rather than VQA questions, and ARO and GQA include non-spatial relations. Our approach mainly relies on keyword filtering and manual review to ensure the sampled data contains only spatial relations. ARO features image captions with clear lexical structures: \textit{<Object> is <Relation> <Object>.} This structure enables us to easily reconstruct the captions and convert them into VQA format.

\subsection{Analysis of Prompting Techniques}
\label{sec:appendix1}
\begin{table*}[h]
\centering
\scalebox{0.79}{
    \begin{tabular}{l c c c c c c c c c}
    \toprule
     & & \multicolumn{2}{c}{\textbf{ARO}} & \multicolumn{2}{c}{\textbf{GQA}} & \multicolumn{2}{c}{\textbf{MMRel}} & \multicolumn{2}{c}{\textbf{Average}}\\
    \textbf{Methods} & \textbf{CoT} & \textbf{Acc} & \textbf{F1} & \textbf{Acc} & \textbf{F1} & \textbf{Acc} & \textbf{F1} & \textbf{Acc} & \textbf{F1}\\
    \midrule
    \multirow{2}{*}{Bidirectional} & No & \textbf{76.00} & \textbf{79.07} & \textbf{70.67} & \textbf{74.27} & 86.83 & 87.32 & \textbf{77.83} & 80.22\\ 
    & Yes & 75.33 & 79.03 & 69.47 & 73.86 & \textbf{88.50} & \textbf{89.12} & 77.77 & \textbf{80.67}\\  
    \midrule
    \multirow{2}{*}{Transitivity} & No & 70.67 & 70.57 & 67.17 & 66.21 & 76.17 & 72.12 & 71.34	& 69.63\\ 
    & Yes & \textbf{73.90} & \textbf{75.89} & \textbf{68.93} & \textbf{71.04} & \textbf{84.03} & \textbf{83.19} & \textbf{75.62} & \textbf{76.71}\\ 
    \midrule
    \multirow{2}{*}{Combined} & No & 71.50 & 76.08 & 66.00 & 71.90 & 83.50 & 84.75 & 73.67 & 77.58 \\
    & Yes & \textbf{76.67} & \textbf{79.57} & \textbf{70.77} & \textbf{75.08} & \textbf{92.70} & \textbf{92.91} & \textbf{80.05} & \textbf{82.52}\\ 
     
    \bottomrule
    \end{tabular}
}\caption{The comparison of results with or without using CoT prompting in  bidirectional, transitivity, and combined constraints.}
\label{analysis_cot}
\end{table*}

\begin{table*}[h]
\centering
\scalebox{0.79}{
    \begin{tabular}{l c c c c c c c c c}
    \toprule
     & & \multicolumn{2}{c}{\textbf{ARO}} & \multicolumn{2}{c}{\textbf{GQA}} & \multicolumn{2}{c}{\textbf{MMRel}} & \multicolumn{2}{c}{\textbf{Average}}\\
    \textbf{Methods} & \textbf{Structure} & \textbf{Acc} & \textbf{F1} & \textbf{Acc} & \textbf{F1} & \textbf{Acc} & \textbf{F1} & \textbf{Acc} & \textbf{F1}\\
    \midrule
    \multirow{2}{*}{Bidirectional} & No & 71.5 & 71.83 & 67.17 & 66.67 & 74.67 & 69.72 & 71.11	& 69.41\\ 
    & Yes & \textbf{75.33} & \textbf{79.03} & \textbf{69.47} & \textbf{73.86} & \textbf{88.50} & \textbf{89.12} & \textbf{77.77} & \textbf{80.67}\\ 
    \midrule
    \multirow{2}{*}{Transitivity} & No & 70.33 & 73.43 & 64.17 & 65.71 & 76.50 & 76.38 & 70.33	& 71.84\\ 
    & Yes & \textbf{73.90} & \textbf{75.89} & \textbf{68.93} & \textbf{71.04} & \textbf{84.03} & \textbf{83.19} & \textbf{75.62} & \textbf{76.71}\\ 
    \midrule
    \multirow{2}{*}{Combined} & No & 75.33 & 78.43 & \textbf{71.67} & 75.07 & 87.17 & 87.64 & 78.06 & 80.38\\
    & Yes & \textbf{76.67} & \textbf{79.57} & 70.77 & \textbf{75.08} & \textbf{92.70} & \textbf{92.91} & \textbf{80.05} & \textbf{82.52}\\ 
     
    \bottomrule
    \end{tabular}
}\caption{The comparison of results with or without using structured reasoning output in bidirectional, transitivity, and combined constraints.}
\label{analysis_structure}
\end{table*}

In this section, we present experiments that demonstrate the effectiveness of the prompting techniques discussed in Section~\ref{sec:skeleton}. We do not assert that these prompting techniques alone will significantly improve the accuracy of LVLMs on spatial relation tasks. Instead, we highlight how our constraint-aware methods can effectively integrate these techniques to enhance overall performance.

\paragraph{CoT}
\noindent In this analysis, we compare GPT-4o's performance using bidirectional, transitivity, and combined constraints, both with and without CoT prompting. To control for CoT, we either include or omit the phrase "think step by step" in the prompt. This analysis follows the same dataset and model settings outlined in Section~\ref{sec:settings}.

As shown in Table~\ref{analysis_cot}, the accuracy and F1 score when using CoT prompting are consistently higher than without CoT for both transitivity and combined methods. Although in some cases, bidirectional methods without CoT outperform those with CoT, the overall performance with CoT remains superior. Thus, we think using CoT can bring benefits to our methods and increases the reasoning ability of LVLMs in general.

\paragraph{Reasoning Structure}
\noindent Similarly, we evaluate the effectiveness of the reasoning structure by comparing the performance of GPT-4o with and without this structure. To create a scenario without the reasoning structure, we remove the output format from the prompt and provide additional descriptive instructions that convey the information originally included in the structured reasoning output, such as analyzing horizontal, vertical, and depth relations.

The results are shown in Table~\ref{analysis_structure}. It is evident that explicitly stating the reasoning requirements in the output format enhances the effectiveness of our method across all three datasets and methods. We hypothesize that LVLMs adhere more strictly to the output format than to descriptive instructions. Consequently, including requirements in the output format increases the likelihood that LVLMs will follow these guidelines.

\subsection{Tradeoff between Accuracy and Cost}
\label{sec:tradeoff_analysis}
\begin{table*}[h]
\centering
\scalebox{0.79}{
    \begin{tabular}{l c c c c c}
    \toprule
     & vanilla & CoT+structure & bidirectional & transitivity & combined\\
    \midrule
    Cost per 100 Questions (\$) & 0.242	& 0.377	& 0.557	& 0.630	& 0.765\\  
    \midrule
    Average Accuracy (\%) & 66.37 & 71.02 & 77.77 & 75.62 & 80.05\\    
    \bottomrule
    \end{tabular}
}\caption{The results of analyzing the tradeoff between accuracy and API cost.}
\label{tab:cost_tab}
\end{table*}
Compared to the vanilla prompt and the prompt incorporating CoT and reasoning structure, the prompts used in our method are longer. While they improve the accuracy of model performance on tasks, they also increase the API cost. Therefore, it is crucial to analyze the accuracy improvement relative to the increase in cost. To investigate this, we design an experiment in which we apply each prompting method to 1,800 questions across three datasets and calculate the average GPT-4o API cost per 100 VQA questions for each method. The results are shown in Table~\ref{tab:cost_tab}. 

From these results, we observe that the cost of our methods is approximately two to three times higher than that of the vanilla prompt. However, the bidirectional method proves to be the most cost-efficient. Therefore, when the budget is limited, users can opt for the bidirectional method to save costs while maintaining performance. When the budget allows, users can select the transitivity or combined method, depending on their preference, to achieve the best accuracy.

\subsection{Results in Other LVLMs}
\label{sec:additional_appendix}
\begin{table*}[h]
\centering
\scalebox{0.72}{
    \begin{tabular}{cc} 
        \begin{subtable}[h]{0.65\linewidth}
            \centering
            \begin{tabular}{l c c c c}
                \toprule
                \textbf{Methods} & \textbf{Acc} & \textbf{Precision} & \textbf{Recall} & \textbf{F1} \\
                \midrule
                Baseline (vanilla) & 69.50 & 62.97 & \textbf{94.67} & 75.63 \\ 
                Baseline (CoT+structure) & 71.50 & 66.17 & 88.00 & 75.54 \\ 
                Bidirectional (ours) & 70.17 & 63.97 & 92.33 & 75.58 \\ 
                Transitivity (ours) & 73.33 & 66.51 & 94.00 & 77.90 \\ 
                Combined (ours) & \textbf{74.50} & \textbf{68.24} & 91.67 & \textbf{78.24} \\ 
                \bottomrule
            \end{tabular}
            \caption{ARO}
        \end{subtable}
        \hspace{0.5cm} 
        
        \begin{subtable}[h]{0.65\linewidth}
            \centering
            \begin{tabular}{l c c c c}
                \toprule
                \textbf{Methods} & \textbf{Acc} & \textbf{Precision} & \textbf{Recall} & \textbf{F1} \\
                \midrule
                Baseline (vanilla) & 66.17 & 61.41 & 87.00 & 72.00 \\ 
                Baseline (CoT+structure) & 68.50 & 62.25 & \textbf{94.00} & 74.90 \\ 
                Bidirectional (ours) & 70.17 & 63.72 & 93.67 & \textbf{75.84} \\ 
                Transitivity (ours) & 70.50 & 65.04 & 88.67 & 75.04 \\ 
                Combined (ours) & \textbf{72.00} & \textbf{67.28} & 85.67 & 75.37 \\ 
                \bottomrule
            \end{tabular}
            \caption{GQA}
        \end{subtable} \\ \\

        \begin{subtable}[h]{0.65\linewidth}
            \centering
            \begin{tabular}{l c c c c}
                \toprule
                \textbf{Methods} & \textbf{Acc} & \textbf{Precision} & \textbf{Recall} & \textbf{F1} \\
                \midrule
                Baseline (vanilla) & 61.17 & 57.11 & 89.67 & 69.78 \\ 
                Baseline (CoT+structure) & 74.17 & 70.54 & 83.00 & 76.26 \\ 
                Bidirectional (ours) & 82.50 & 77.94 & 90.67 & 83.82 \\ 
                Transitivity (ours) & 82.83 & 76.99 & \textbf{93.67} & 84.51 \\ 
                Combined (ours) & \textbf{85.00} & \textbf{80.88} & 91.67 & \textbf{85.94} \\ 
                \bottomrule
            \end{tabular}
            \caption{MMRel}
        \end{subtable}
        \hspace{0.5cm} 
            
        \begin{subtable}[h]{0.65\linewidth}
            \centering
            \begin{tabular}{l c c c c}
                \toprule
                \textbf{Methods} & \textbf{Acc} & \textbf{Precision} & \textbf{Recall} & \textbf{F1} \\
                \midrule
                Baseline (vanilla) & 65.61 & 60.50 & 90.45 & 72.47 \\ 
                Baseline (CoT+structure) & 71.39 & 66.32 & 88.33 & 75.57 \\ 
                Bidirectional (ours) & 74.28 & 68.54 & \textbf{92.22} & 78.41 \\ 
                Transitivity (ours) & 75.55 & 69.51 & 92.11 & 79.15 \\ 
                Combined (ours) & \textbf{77.17} & \textbf{72.13} & 89.67 & \textbf{79.85} \\ 
                \bottomrule
            \end{tabular}
            \caption{Average}
        \end{subtable}
    \end{tabular}
}
\caption{Full results of our methods (bidirectional, transitivity, and combined constraints) on three datasets using Gemini Pro.}
\label{tab:add_exp}
\end{table*}

In addition to GPT-4o, we also use other models to evaluate our methods. Under the same experimental settings, we use Gemini Pro\footnote{gemini-1.5-pro-002}~\citep{team2024gemini} to evaluate two baseline methods and three proposed methods. The results are shown in Table~\ref{tab:add_exp}. The performance of our methods is consistently strong in Gemini Pro: the combined method outperforms the bidirectional and transitivity methods, and the three proposed methods are generally better than the two baselines. An exception is the ARO dataset, where bidirectional constraints have slightly lower accuracy than the baseline method incorporating CoT and reasoning structure.

Additionally, we test our methods on smaller (1B or 7B) open-source models, such as LLaVA~\citep{liu2024visual} and Janus Pro~\citep{chen2025janus}. We observe that our methods show potential in mitigating hallucination in these models. However, due to limitations caused by the smaller model sizes, the responses from these models are unstable. In some cases, the models fail to follow the instructions or formats written in the prompt to generate a response.

\subsection{Extended Results}
\label{sec:appendix2}
\begin{table*}[h]
\centering
\scalebox{0.72}{
    \begin{tabular}{cc} 
        \begin{subtable}[h]{0.65\linewidth}
            \centering
            \begin{tabular}{l c c c c}
                \toprule
                \textbf{Methods} & \textbf{Acc} & \textbf{Precision} & \textbf{Recall} & \textbf{F1} \\
                \midrule
                Baseline (vanilla) & 65.10 & 60.90 & 84.40 & 70.75 \\ 
                Baseline (CoT+structure) & 69.60 & 64.15 & 88.87 & 74.51 \\ 
                Bidirectional (ours) & 75.33 & 68.74 & \textbf{92.93} & 79.03 \\ 
                Transitivity (ours) & 73.90 & 70.50 & 82.20 & 75.89 \\ 
                Combined (ours) & \textbf{76.67} & \textbf{70.77} & 90.87 & \textbf{79.57} \\ 
                \bottomrule
            \end{tabular}
            \caption{ARO}
        \end{subtable}
        \hspace{0.5cm} 
        
        \begin{subtable}[h]{0.65\linewidth}
            \centering
            \begin{tabular}{l c c c c}
                \toprule
                \textbf{Methods} & \textbf{Acc} & \textbf{Precision} & \textbf{Recall} & \textbf{F1} \\
                \midrule
                Baseline (vanilla) & 63.03 & 60.09 & 77.60 & 67.73 \\ 
                Baseline (CoT+structure) & 63.47 & 60.48 & 77.73 & 68.03 \\ 
                Bidirectional (ours) & 69.47 & 64.57 & 86.27 & 73.86 \\ 
                Transitivity (ours) & 68.93 & \textbf{66.54} & 76.20 & 71.04 \\ 
                Combined (ours) & \textbf{70.77} & 65.43 & \textbf{88.07} & \textbf{75.08} \\ 
                \bottomrule
            \end{tabular}
            \caption{GQA}
        \end{subtable} \\ \\

        \begin{subtable}[h]{0.65\linewidth}
            \centering
            \begin{tabular}{l c c c c}
                \toprule
                \textbf{Methods} & \textbf{Acc} & \textbf{Precision} & \textbf{Recall} & \textbf{F1} \\
                \midrule
                Baseline (vanilla) & 70.96 & 65.72 & 87.67 & 75.12 \\ 
                Baseline (CoT+structure) & 80.00 & 76.04 & 87.60 & 81.41 \\ 
                Bidirectional (ours) & 88.50 & 84.57 & 94.20 & 89.12 \\ 
                Transitivity (ours) & 84.03 & 87.80 & 79.07 & 83.19 \\ 
                Combined (ours) & \textbf{92.70} & \textbf{90.37} & \textbf{95.60} & \textbf{92.91} \\ 
                \bottomrule
            \end{tabular}
            \caption{MMRel}
        \end{subtable}
        \hspace{0.5cm} 
            
        \begin{subtable}[h]{0.65\linewidth}
            \centering
            \begin{tabular}{l c c c c}
                \toprule
                \textbf{Methods} & \textbf{Acc} & \textbf{Precision} & \textbf{Recall} & \textbf{F1} \\
                \midrule
                Baseline (vanilla) & 66.37 & 62.24 & 83.22 & 71.20 \\ 
                Baseline (CoT+structure) & 71.02 & 66.89 & 84.73 & 74.65 \\ 
                Bidirectional (ours) & 77.77 & 72.63 & 91.13 & 80.67 \\ 
                Transitivity (ours) & 75.62 & 74.95 & 79.16 & 76.71 \\ 
                Combined (ours) & \textbf{80.05} & \textbf{75.52} & \textbf{91.51} & \textbf{82.52} \\ 
                \bottomrule
            \end{tabular}
            \caption{Average}
        \end{subtable}
    \end{tabular}
}
\caption{Full results of our methods (bidirectional, transitivity, and combined constraints) on three datasets using GPT-4o. This is the extended data of Table~\ref{tab:main_results}.}
\label{tab:main_results_full}
\end{table*}

\begin{figure*}[h]
    \centering
    \includegraphics[width=\textwidth]{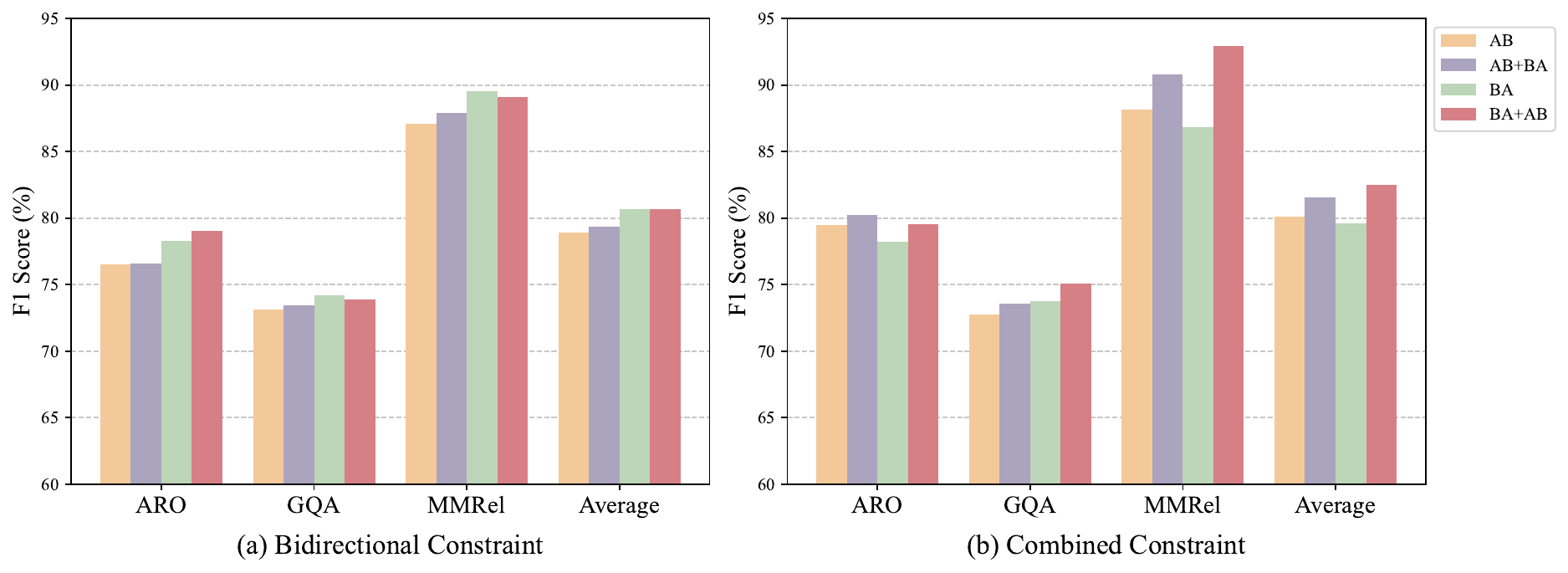}
    \caption{The F1 score comparison of different relation analysis choices in bidirectional and combined constraints is shown.}
    \label{fig:relation_f1}
\end{figure*}

\begin{table*}[h]
\centering
\scalebox{0.72}{
    \begin{tabular}{cc}
        \begin{subtable}[h]{0.65\linewidth}
            \centering
            \begin{tabular}{l c c c c}
                \toprule
                \textbf{Relations} & \textbf{Acc} & \textbf{Precision} & \textbf{Recall} & \textbf{F1} \\
                \midrule
                AB & 71.70 & 65.38 & 92.27 & 76.53 \\ 
                BA & 74.67 & 68.54 & 91.20 & 78.26 \\ 
                AB+BA & 71.77 & 65.38 & 92.53 & 76.62 \\ 
                BA+AB & \textbf{75.33} & \textbf{68.74} & \textbf{92.93} & \textbf{79.03} \\ 
                \bottomrule
            \end{tabular}
            \caption{ARO}
        \end{subtable}
        \hspace{-2cm}
        
        \begin{subtable}[h]{0.65\linewidth}
            \centering
            \begin{tabular}{l c c c c}
                \toprule
                \textbf{Relations} & \textbf{Acc} & \textbf{Precision} & \textbf{Recall} & \textbf{F1} \\
                \midrule
                AB & 68.03 & 63.09 & 86.93 & 73.11 \\ 
                BA & \textbf{69.87} & \textbf{64.84} & 86.80 & \textbf{74.23} \\ 
                AB+BA & 68.07 & 62.85 & \textbf{88.22} & 73.46 \\ 
                BA+AB & 69.47 & 64.57 & 86.27 & 73.86 \\ 
                \bottomrule
            \end{tabular}
            \caption{GQA}
        \end{subtable} \\ \\

        \begin{subtable}[h]{0.65\linewidth}
            \centering
            \begin{tabular}{l c c c c}
                \toprule
                \textbf{Relations} & \textbf{Acc} & \textbf{Precision} & \textbf{Recall} & \textbf{F1} \\
                \midrule
                AB & 86.50 & 82.97 & 91.87 & 87.10 \\ 
                BA & \textbf{89.10} & \textbf{85.97} & 93.47 & \textbf{89.56} \\ 
                AB+BA & 87.20 & 83.22 & 93.20 & 87.92 \\ 
                BA+AB & 88.50 & 84.57 & \textbf{94.20} & 89.12 \\ 
                \bottomrule
            \end{tabular}
            \caption{MMRel}
        \end{subtable}
        \hspace{-2cm}
        
        \begin{subtable}[h]{0.65\linewidth}
            \centering
            \begin{tabular}{l c c c c}
                \toprule
                \textbf{Relations} & \textbf{Acc} & \textbf{Precision} & \textbf{Recall} & \textbf{F1} \\
                \midrule
                AB & 75.41 & 70.48 & 90.36 & 78.91 \\ 
                BA & \textbf{77.88} & \textbf{73.12} & 90.49 & \textbf{80.68} \\ 
                AB+BA & 75.68 & 70.48 & \textbf{91.32} & 79.33 \\ 
                BA+AB & 77.77 & 72.63 & 91.13 & 80.67 \\ 
                \bottomrule
            \end{tabular}
            \caption{Average}
        \end{subtable}
    \end{tabular}
}
\caption{Full results of the comparison between different relation choices in the bidirectional constraints.}
\label{tab:bidirection_relations_full}
\end{table*}

\begin{table*}[h]
\centering
\scalebox{0.72}{
    \begin{tabular}{cc} 
        \begin{subtable}[h]{0.65\linewidth}
            \centering
            \begin{tabular}{l c c c c}
                \toprule
                \textbf{Relations} & \textbf{Acc} & \textbf{Precision} & \textbf{Recall} & \textbf{F1} \\
                \midrule
                AB & 77.00 & \textbf{71.77} & 89.00 & 79.46 \\ 
                BA & 75.83 & 71.23 & 86.67 & 78.20 \\ 
                AB+BA & \textbf{77.53} & 71.60 & \textbf{91.27} & \textbf{80.25} \\ 
                BA+AB & 76.67 & 70.77 & 90.87 & 79.57 \\ 
                \bottomrule
            \end{tabular}
            \caption{ARO}
        \end{subtable}
        \hspace{-2cm} 
        
        \begin{subtable}[h]{0.65\linewidth}
            \centering
            \begin{tabular}{l c c c c}
                \toprule
                \textbf{Relations} & \textbf{Acc} & \textbf{Precision} & \textbf{Recall} & \textbf{F1} \\
                \midrule
                AB & 68.17 & 63.59 & 85.00 & 72.75 \\ 
                BA & 70.67 & \textbf{66.76} & 82.33 & 73.73 \\ 
                AB+BA & 69.03 & 64.15 & 86.33 & 73.60 \\ 
                BA+AB & \textbf{70.77} & 65.43 & \textbf{88.07} & \textbf{75.08} \\ 
                \bottomrule
            \end{tabular}
            \caption{GQA}
        \end{subtable} \\ \\

        \begin{subtable}[h]{0.65\linewidth}
            \centering
            \begin{tabular}{l c c c c}
                \toprule
                \textbf{Relations} & \textbf{Acc} & \textbf{Precision} & \textbf{Recall} & \textbf{F1} \\
                \midrule
                AB & 88.50 & 90.81 & 85.67 & 88.16 \\ 
                BA & 87.17 & 89.12 & 84.67 & 86.84 \\ 
                AB+BA & 90.83 & \textbf{90.97} & 90.67 & 90.82 \\ 
                BA+AB & \textbf{92.70} & 90.37 & \textbf{95.60} & \textbf{92.91} \\ 
                \bottomrule
            \end{tabular}
            \caption{MMRel}
        \end{subtable}
        \hspace{-2cm} 
            
        \begin{subtable}[h]{0.65\linewidth}
            \centering
            \begin{tabular}{l c c c c}
                \toprule
                \textbf{Relations} & \textbf{Acc} & \textbf{Precision} & \textbf{Recall} & \textbf{F1} \\
                \midrule
                AB & 77.89 & 75.39 & 86.56 & 80.12 \\ 
                BA & 77.89 & \textbf{75.70} & 84.56 & 79.59 \\ 
                AB+BA & 79.13 & 75.57 & 89.42 & 81.56 \\ 
                BA+AB & \textbf{80.05} & 75.52 & \textbf{91.51} & \textbf{82.52} \\ 
                \bottomrule
            \end{tabular}
            \caption{Average}
        \end{subtable}
    \end{tabular}
}
\caption{Full results of the comparison between different relation choices in the combined constraints.}
\label{tab:combined_relations_full}
\end{table*}

\begin{table*}[h]
\centering
\scalebox{0.72}{
    \begin{tabular}{cc} 
        \begin{subtable}[h]{0.65\linewidth}
            \centering
            \begin{tabular}{l c c c c}
                \toprule
                \textbf{Attributes} & \textbf{Acc} & \textbf{Precision} & \textbf{Recall} & \textbf{F1} \\
                \midrule
                The largest & \textbf{75.00} & 71.43 & 83.33 & 76.92 \\ 
                The smallest & 73.33 & 69.34 & 83.67 & 75.83 \\ 
                The most top & 72.17 & 69.97 & 77.67 & 73.62 \\ 
                The central & \textbf{75.00} & \textbf{71.68} & 82.67 & 76.78 \\ 
                The most obvious & \textbf{75.00} & 71.19 & \textbf{84.00} & \textbf{77.06} \\ 
                Random & 73.90 & 70.50 & 82.20 & 75.89 \\ 
                \bottomrule
            \end{tabular}
            \caption{ARO}
        \end{subtable}
        \hspace{-0.7cm} 
        
        \begin{subtable}[h]{0.65\linewidth}
            \centering
            \begin{tabular}{l c c c c}
                \toprule
                \textbf{Attributes} & \textbf{Acc} & \textbf{Precision} & \textbf{Recall} & \textbf{F1} \\
                \midrule
                The largest & 65.00 & 63.01 & 72.67 & 67.49 \\ 
                The smallest & 66.00 & 64.29 & 72.00 & 67.92 \\ 
                The most top & 65.50 & 63.48 & 73.00 & 67.91 \\ 
                The central & \textbf{69.17} & \textbf{67.27} & 74.67 & 70.77 \\ 
                The most obvious & 67.33 & 65.29 & 74.00 & 69.38 \\ 
                Random & 68.93 & 66.54 & \textbf{76.20} & \textbf{71.04} \\ 
                \bottomrule
            \end{tabular}
            \caption{GQA}
        \end{subtable} \\ \\

        \begin{subtable}[h]{0.65\linewidth}
            \centering
            \begin{tabular}{l c c c c}
                \toprule
                \textbf{Attributes} & \textbf{Acc} & \textbf{Precision} & \textbf{Recall} & \textbf{F1} \\
                \midrule
                The largest & 83.83 & 86.38 & \textbf{80.33} & \textbf{83.25} \\ 
                The smallest & 84.00 & \textbf{88.64} & 78.00 & 82.98 \\ 
                The most top & 81.67 & 84.67 & 77.33 & 80.84 \\ 
                The central & 83.00 & 88.37 & 76.00 & 81.72 \\ 
                The most obvious & 81.83 & 86.04 & 76.00 & 80.71 \\ 
                Random & \textbf{84.03} & 87.80 & 79.07 & 83.19 \\ 
                \bottomrule
            \end{tabular}
            \caption{MMRel}
        \end{subtable}
        \hspace{-0.7cm} 
            
        \begin{subtable}[h]{0.65\linewidth}
            \centering
            \begin{tabular}{l c c c c}
                \toprule
                \textbf{Attributes} & \textbf{Acc} & \textbf{Precision} & \textbf{Recall} & \textbf{F1} \\
                \midrule
                \text{The largest} & 74.61 & 73.61 & 78.78 & 75.89 \\ 
                \text{The smallest} & 74.44 & 74.76 & 77.22 & 75.58 \\ 
                \text{The most top} & 73.11 & 72.71 & 76.00 & 74.12 \\ 
                \text{The central} & \textbf{75.72} & \textbf{75.77} & 77.78 & 76.42 \\ 
                \text{The most obvious} & 74.72 & 74.84 & 78.00 & 75.72 \\ 
                \text{Random} & 75.62 & 74.95 & \textbf{79.16} & \textbf{76.71} \\ 
                \bottomrule
            \end{tabular}
            \caption{Average}
        \end{subtable}
    \end{tabular}
}
\caption{Full results of the comparison of different reference object selection strategies in transitivity constraints. This is the extended data of Table~\ref{tab:analysis1}.}
\label{tab:analysis1_full}
\end{table*}

\begin{table*}[h]
\centering
\scalebox{0.72}{
    \begin{tabular}{cc} 
        \begin{subtable}[h]{0.65\linewidth}
            \centering
            \begin{tabular}{l c c c c}
                \toprule
                \textbf{Attributes} & \textbf{Acc} & \textbf{Precision} & \textbf{Recall} & \textbf{F1} \\
                \midrule
                The largest & 77.00 & 71.66 & 89.33 & 79.53 \\ 
                The smallest & 75.83 & 69.82 & 91.00 & 79.02 \\ 
                The most top & 76.17 & 70.18 & 91.00 & 79.25 \\ 
                The central & \textbf{78.17} & \textbf{72.06} & \textbf{92.00} & \textbf{80.82} \\ 
                The most obvious & 77.67 & 71.61 & 91.67 & 80.41 \\ 
                Random & 76.67 & 70.77 & 90.87 & 79.57 \\ 
                \bottomrule
            \end{tabular}
            \caption{ARO}
        \end{subtable}
        \hspace{-0.7cm} 
        
        \begin{subtable}[h]{0.65\linewidth}
            \centering
            \begin{tabular}{l c c c c}
                \toprule
                \textbf{Attributes} & \textbf{Acc} & \textbf{Precision} & \textbf{Recall} & \textbf{F1} \\
                \midrule
                The largest & 70.50 & 65.04 & \textbf{88.67} & 75.04 \\ 
                The smallest & 69.67 & 64.39 & 88.00 & 74.37 \\ 
                The most top & 68.83 & 63.95 & 86.33 & 73.48 \\ 
                The central & 68.83 & 63.81 & 87.00 & 73.62 \\ 
                The most obvious & 69.67 & 64.82 & 86.00 & 73.93 \\ 
                Random & \textbf{70.77} & \textbf{65.43} & 88.07 & \textbf{75.08} \\ 
                \bottomrule
            \end{tabular}
            \caption{GQA}
        \end{subtable} \\ \\

        \begin{subtable}[h]{0.65\linewidth}
            \centering
            \begin{tabular}{l c c c c}
                \toprule
                \textbf{Attributes} & \textbf{Acc} & \textbf{Precision} & \textbf{Recall} & \textbf{F1} \\
                \midrule
                The largest & 91.67 & 89.56 & 94.33 & 91.88 \\ 
                The smallest & 91.17 & 88.96 & 94.00 & 91.41 \\ 
                The most top & 92.33 & 89.44 & \textbf{96.00} & 92.60 \\ 
                The central & 90.67 & 88.61 & 93.33 & 90.91 \\ 
                The most obvious & 92.33 & \textbf{90.71} & 94.33 & 92.48 \\ 
                Random & \textbf{92.70} & 90.37	& 95.60 & \textbf{92.91} \\ 
                \bottomrule
            \end{tabular}
            \caption{MMRel}
        \end{subtable}
        \hspace{-0.7cm} 
            
        \begin{subtable}[h]{0.65\linewidth}
            \centering
            \begin{tabular}{l c c c c}
                \toprule
                \textbf{Attributes} & \textbf{Acc} & \textbf{Precision} & \textbf{Recall} & \textbf{F1} \\
                \midrule
                \text{The largest} & 79.72 & 75.42 & 90.78 & 82.15 \\ 
                \text{The smallest} & 78.89 & 74.39 & 91.00 & 81.60 \\ 
                \text{The most top} & 79.11 & 74.52 & 91.11 & 81.78 \\ 
                \text{The central} & 79.22 & 74.16 & 90.78 & 81.78 \\ 
                \text{The most obvious} & 79.89 & 75.05 & 90.67 & 82.27 \\ 
                \text{Random} & \textbf{80.05} & \textbf{75.52} & \textbf{91.51} & \textbf{82.52} \\ 
                \bottomrule
            \end{tabular}
            \caption{Average}
        \end{subtable}
    \end{tabular}
}
\caption{Full results of the comparison of different reference object selection strategies in combined constraints. This is the extended data of Table~\ref{tab:analysis2}.}
\label{tab:analysis2_full}
\end{table*}

This section presents the full results, including accuracy, precision, recall, and F1 score, for all experiments conducted in the body of the paper.

The comprehensive results of the main experiment, which include a comparison of our methods against baseline approaches across four metrics, are presented in Table~\ref{tab:main_results_full}. In addition to evaluating the accuracy of various bidirectional relation analysis options, as discussed in Section~\ref{sec:analysis1}, we also assess the F1 scores for different analytical choices, as shown in Figure~\ref{fig:relation_f1}. The F1 score comparison adheres to the same criteria as the accuracy evaluations, and the results follow similar patterns. Detailed data on the relation analysis options can be found in Tables~\ref{tab:bidirection_relations_full} and~\ref{tab:combined_relations_full}. Furthermore, extended results on reference object selection analysis is available in Tables~\ref{tab:analysis1_full} and~\ref{tab:analysis2_full}.

\subsection{Failed Case Analysis}
\label{sec:appendix6}
\begin{figure*}[h]
    \centering
    \begin{overpic}[width=1\linewidth]{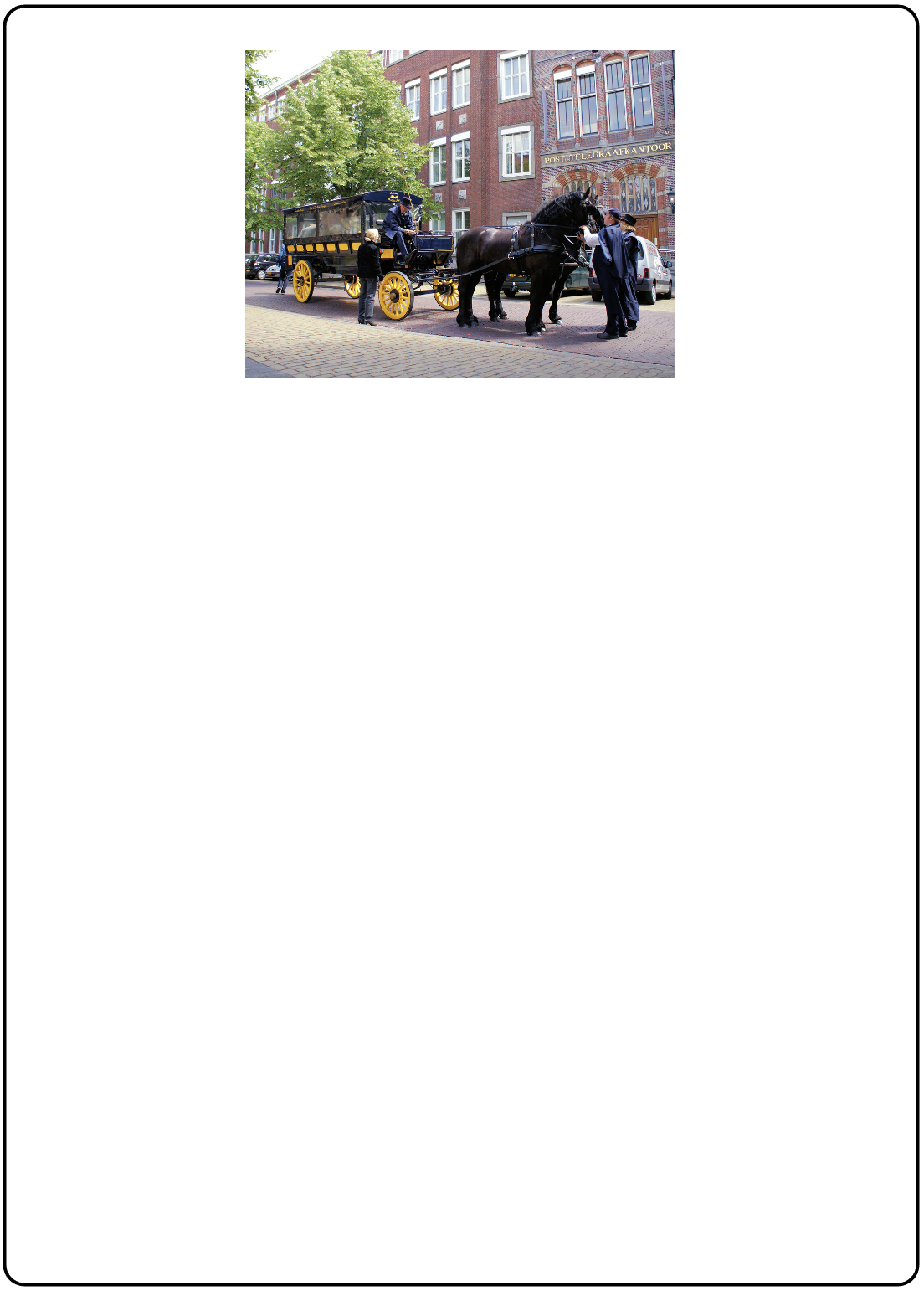}
        \put(3,36){\textcolor{black}{
        \begin{minipage}{0.91\linewidth}
            \fontsize{8}{12}\selectfont
            \textbf{Q:}
            \\\#\# Instructions \#\#
            \\1. Repeat the question and then extract the objects mentioned in the question. Label the first object that appears as "Object A" and the second as "Object B".
            \\2. Describe the relative position between Object B and A and between Object A and B.
            \\3. Reference to the relationship between Object B and A and between Object A and B, and then think step by step to use 'yes' or 'no' to answer the question.
            \\
            \\\#\# Please output in the following format \#\#
            \\Question:
            \\Object A:
            \\Object B:
            \\Horizontal relation between Object B and A: B is <relation> A
            \\Vertical relation between Object B and A: B is <relation> A
            \\Depth relation between Object B and A: B is <relation> A
            \\Horizontal relation between Object A and B: A is <relation> B
            \\Vertical relation between Object A and B: A is <relation> B
            \\Depth relation between Object A and B: A is <relation> B
            \\Thinking process: 
            \\Answer: 
            \\
            \\\#\# Question \#\#
            \\Is there a building to the right of the horses in the image?
            \\
            \\\textbf{A:}
            \\Question: Is there a building to the right of the horses in the image?
            \\Object A: horses
            \\Object B: building
            \\Horizontal relation between Object B and A: B is to the right of A
            \\Vertical relation between Object B and A: B is behind A
            \\Depth relation between Object B and A: B is behind A
            \\Horizontal relation between Object A and B: A is to the left of B
            \\Vertical relation between Object A and B: A is in front of B
            \\Depth relation between Object A and B: A is in front of B
            \\Thinking process: The building is positioned to the right and behind the horses in the image.
            \\Answer: Yes
        \end{minipage}}}
    \end{overpic}
    \caption{Failed case 1.} 
    \label{fig:failed1}
\end{figure*}

\begin{figure*}[h]
    \centering
    \begin{overpic}[width=1\linewidth]{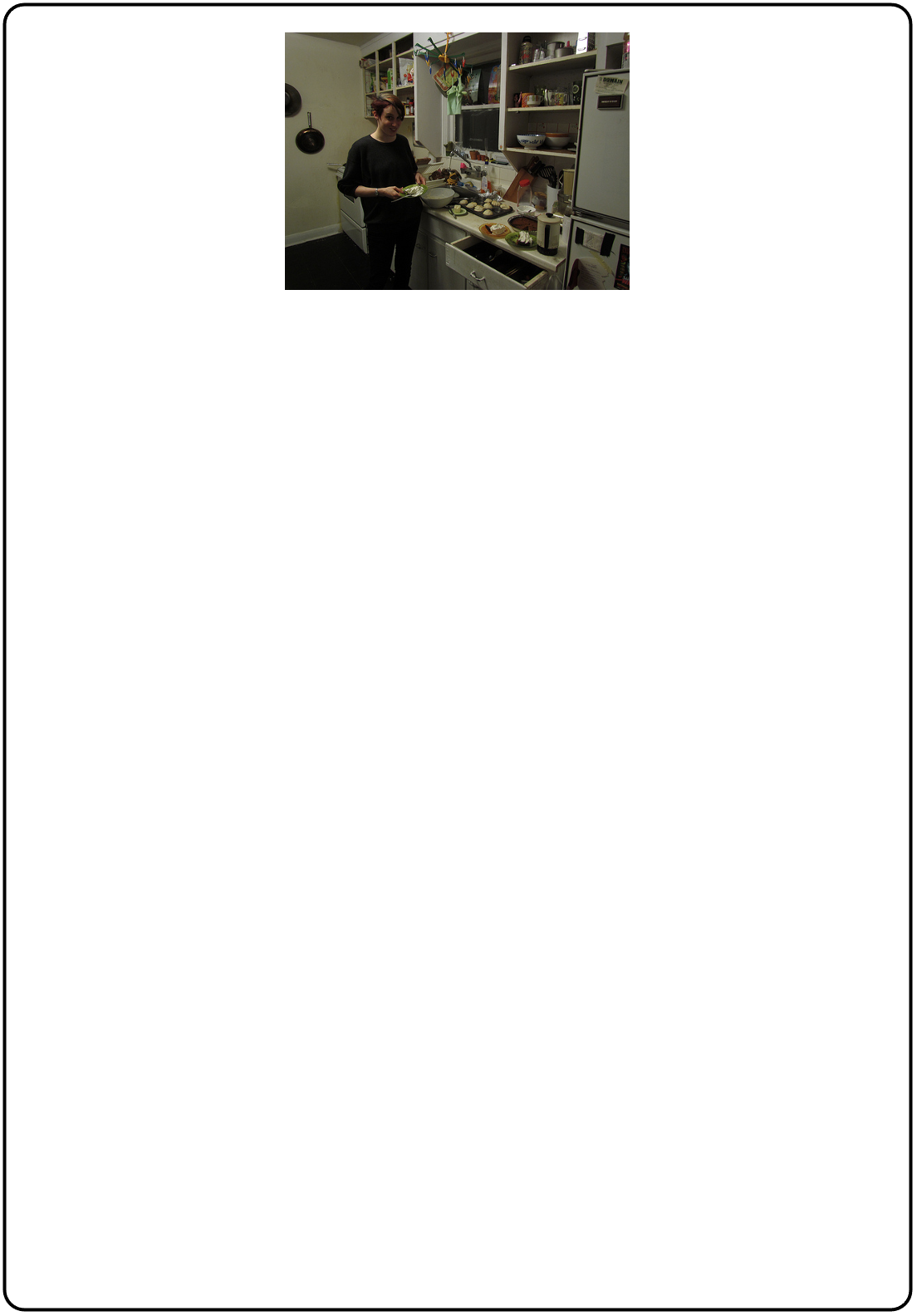}
        \put(3,39){\textcolor{black}{
        \begin{minipage}{0.91\linewidth}
            \fontsize{8}{12}\selectfont
            \textbf{Q:}
            \\\#\# Instructions \#\#
            \\1. Repeat the question and then extract the objects mentioned in the question. Label the first object that appears as "Object A" and the second as "Object B". Select an object different from Object A or Object B in the image as "Object C"
            \\2. Describe the relative position between Object A and C and between Object B and C.
            \\3. Reference to the relationship between Object A and C and between Object B and C, and then think step by step to use 'yes' or 'no' to answer the question.
            \\
            \\\#\# Please output in the following format \#\#
            \\Question:
            \\Object A:
            \\Object B:
            \\Object C:
            \\Horizontal relation between Object A and C: A is <relation> C
            \\Vertical relation between Object A and C: A is <relation> C
            \\Depth relation between Object A and C: A is <relation> C
            \\Horizontal relation between Object B and C: B is <relation> C
            \\Vertical relation between Object B and C: B is <relation> C
            \\Depth relation between Object B and C: B is <relation> C
            \\Thinking process: 
            \\Answer: 
            \\
            \\\#\# Question \#\#
            \\Is there a cabinet to the left of the wall in the image?
            \\
            \\\textbf{A:}
            \\Question: Is there a cabinet to the left of the wall in the image?
            \\Object A: cabinet
            \\Object B: wall
            \\Object C: refrigerator
            \\Horizontal relation between Object A and C: A is to the left of C
            \\Vertical relation between Object A and C: A is above C
            \\Depth relation between Object A and C: A is in front of C
            \\Horizontal relation between Object B and C: B is to the left of C
            \\Vertical relation between Object B and C: B is above C
            \\Depth relation between Object B and C: B is in front of C
            \\Thinking process: The cabinet (Object A) is to the left of the refrigerator (Object C), and the wall (Object B) is also to the left of the refrigerator (Object C). Since both the cabinet and the wall are to the left of the refrigerator, it can be concluded that the cabinet is to the left of the wall.
            \\Answer: Yes
        \end{minipage}}}
    \end{overpic}
    \caption{Failed case 2.} 
    \label{fig:failed2}
\end{figure*}

Although our methods significantly improve the accuracy of LVLMs in reasoning about visual spatial relations, some failure cases persist. Figure~\ref{fig:failed1} illustrates a typical example. In this case, the model uses bidirectional constraints to determine the spatial relation between a building and a horse. The first issue is that GPT fails to label the initially mentioned object, the building, as "Object A," instead switching the labels of the building and the horse. However, this is not the critical error. The main issue lies in GPT's inability to correctly identify the spatial relation from the horse's perspective, incorrectly stating that the building is to the right of the horse. A secondary mistake, though not impacting the final answer, is the use of "front" and "behind" to describe a vertical relationship. This analysis suggests that if the candidate LVLM has inherent misunderstandings or hallucinations regarding the definition of spatial relations, correcting them may be challenging. 

Figure~\ref{fig:failed2} presents a typical failure case involving transitivity constraints. In this instance, GPT applies transitivity constraints to determine the horizontal relationship between a cabinet and a wall, using the refrigerator as a reference object. While the LVLM correctly identifies that "the cabinet is to the left of the refrigerator" and "the wall is to the left of the refrigerator," these spatial relations do not contribute to answering the question accurately. As a result, the final response is still hallucinated.

\subsection{Output Examples}
\label{sec:appendix5}
\begin{figure*}[h]
    \centering
    \begin{overpic}[width=1\linewidth]{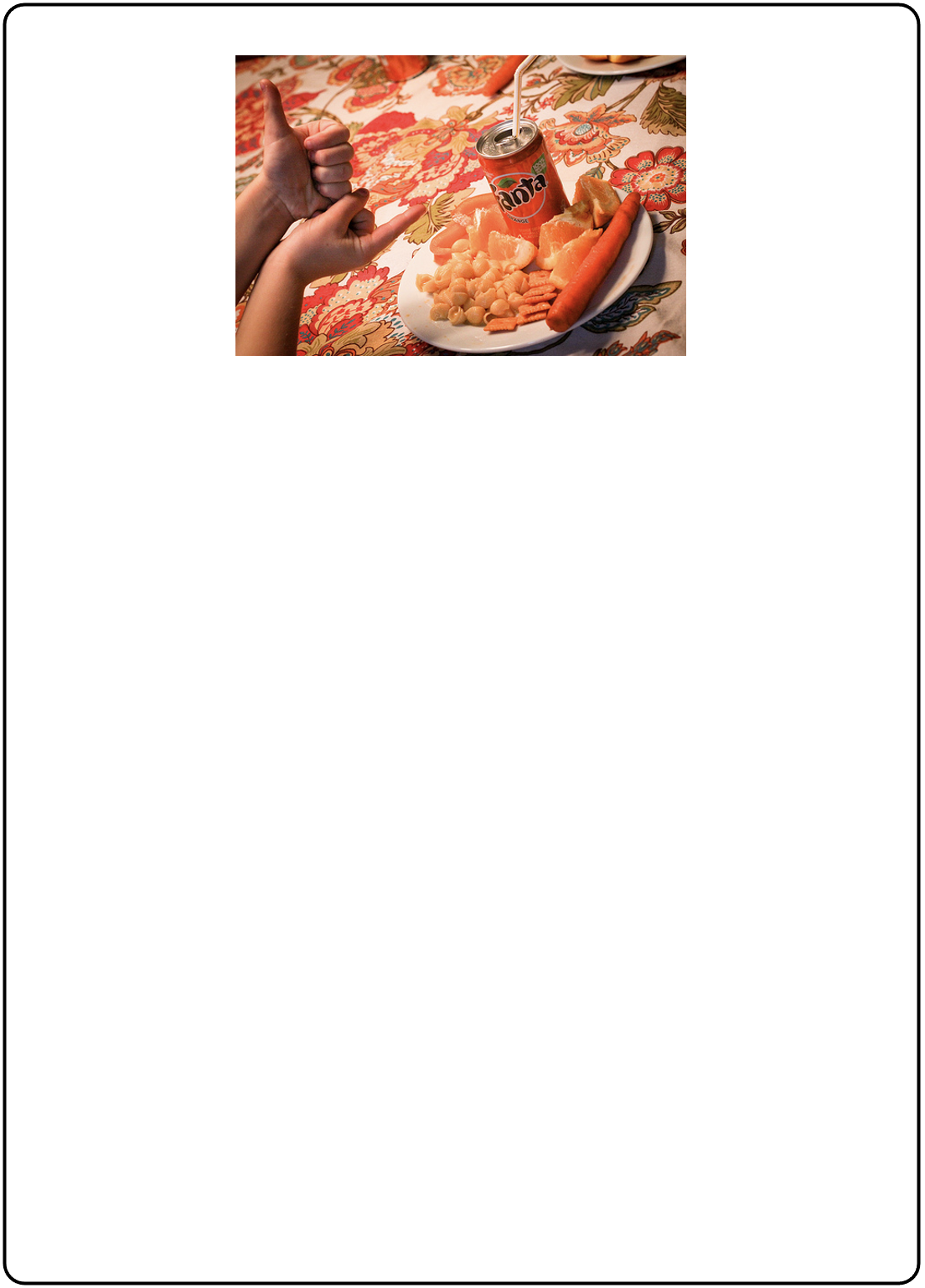}
        \put(3,36){\textcolor{black}{
        \begin{minipage}{0.91\linewidth}
            \fontsize{8}{12}\selectfont
            \textbf{Q:}
            \\\#\# Instructions \#\#
            \\1. Repeat the question and then extract the objects mentioned in the question. Label the first object that appears as "Object A" and the second as "Object B".
            \\2. Describe the relative position between Object B and A and between Object A and B.
            \\3. Reference to the relationship between Object B and A and between Object A and B, and then think step by step to use 'yes' or 'no' to answer the question.
            \\
            \\\#\# Please output in the following format \#\#
            \\Question:
            \\Object A:
            \\Object B:
            \\Horizontal relation between Object B and A: B is <relation> A
            \\Vertical relation between Object B and A: B is <relation> A
            \\Depth relation between Object B and A: B is <relation> A
            \\Horizontal relation between Object A and B: A is <relation> B
            \\Vertical relation between Object A and B: A is <relation> B
            \\Depth relation between Object A and B: A is <relation> B
            \\Thinking process: 
            \\Answer: 
            \\
            \\\#\# Question \#\#
            \\Is there a can left the carrot in the image?
            \\
            \\\textbf{A:}
            \\Question: Is there a can left the carrot in the image?
            \\Object A: can
            \\Object B: carrot
            \\Horizontal relation between Object B and A: B is right of A
            \\Vertical relation between Object B and A: B is at the same level as A
            \\Depth relation between Object B and A: B is in front of A
            \\Horizontal relation between Object A and B: A is left of B
            \\Vertical relation between Object A and B: A is at the same level as B
            \\Depth relation between Object A and B: A is behind B
            \\Thinking process: The carrot is to the right of the can, at the same level, and in front of it. Therefore, the can is to the left of the carrot, at the same level, and behind it.
            \\Answer: Yes
        \end{minipage}}}
    \end{overpic}
    \caption{Example output of the bidirectional constraint.} 
    \label{fig:example1}
\end{figure*}

\begin{figure*}[h]
    \centering
    \begin{overpic}[width=1\linewidth]{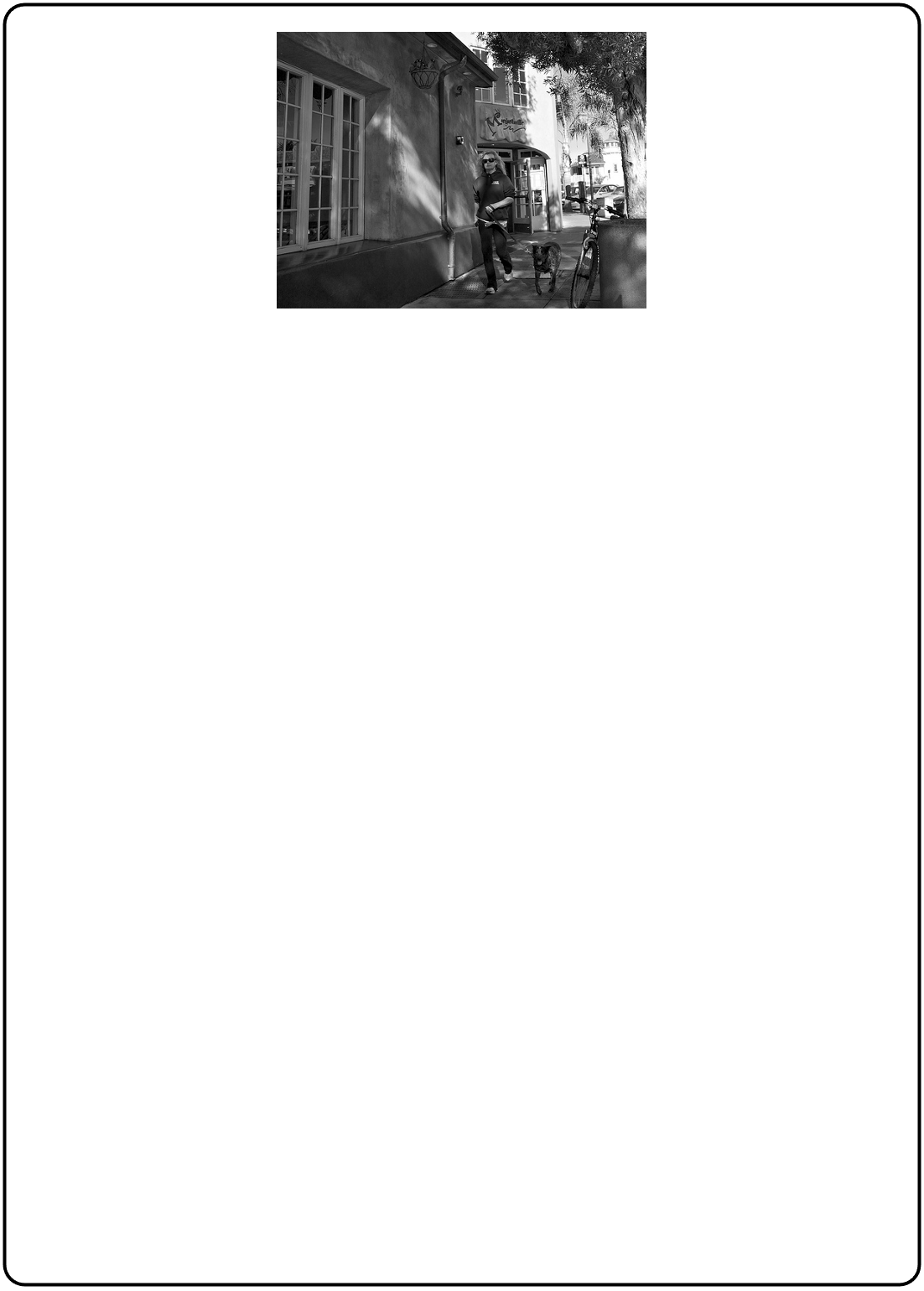}
        \put(3,38){\textcolor{black}{
        \begin{minipage}{0.91\linewidth}
            \fontsize{8}{12}\selectfont
            \textbf{Q:}
            \\\#\# Instructions \#\#
            \\1. Repeat the question and then extract the objects mentioned in the question. Label the first object that appears as "Object A" and the second as "Object B". Select an object different from Object A or Object B in the image as "Object C"
            \\2. Describe the relative position between Object A and C and between Object B and C.
            \\3. Reference to the relationship between Object A and C and between Object B and C, and then think step by step to use 'yes' or 'no' to answer the question.
            \\
            \\\#\# Please output in the following format \#\#
            \\Question:
            \\Object A:
            \\Object B:
            \\Object C:
            \\Horizontal relation between Object A and C: A is <relation> C
            \\Vertical relation between Object A and C: A is <relation> C
            \\Depth relation between Object A and C: A is <relation> C
            \\Horizontal relation between Object B and C: B is <relation> C
            \\Vertical relation between Object B and C: B is <relation> C
            \\Depth relation between Object B and C: B is <relation> C
            \\Thinking process: 
            \\Answer: 
            \\
            \\\#\# Question \#\#
            \\Is there a window to the right of the tree in the image?
            \\
            \\\textbf{A:}
            \\Question: Is there a window to the right of the tree in the image?
            \\Object A: window
            \\Object B: tree
            \\Object C: bicycle
            \\Horizontal relation between Object A and C: A is left of C
            \\Vertical relation between Object A and C: A is above C
            \\Depth relation between Object A and C: A is behind C
            \\Horizontal relation between Object B and C: B is right of C
            \\Vertical relation between Object B and C: B is above C
            \\Depth relation between Object B and C: B is behind C
            \\Thinking process: The window (Object A) is to the left of the bicycle (Object C), and the tree (Object B) is to the right of the bicycle (Object C). Therefore, the window is to the left of the tree.
            \\Answer: No
        \end{minipage}}}
    \end{overpic}
    \caption{Example output of the transitivity constraint.} 
    \label{fig:example2}
\end{figure*}

Figures~\ref{fig:example1} and~\ref{fig:example2} demonstrate the example outputs of bidirectional and transitivity constraints respectively.

\subsection{Template Prompts}
\label{sec:appendix7}
\begin{figure*}[h]
    \centering
    \begin{overpic}[width=0.5\linewidth]{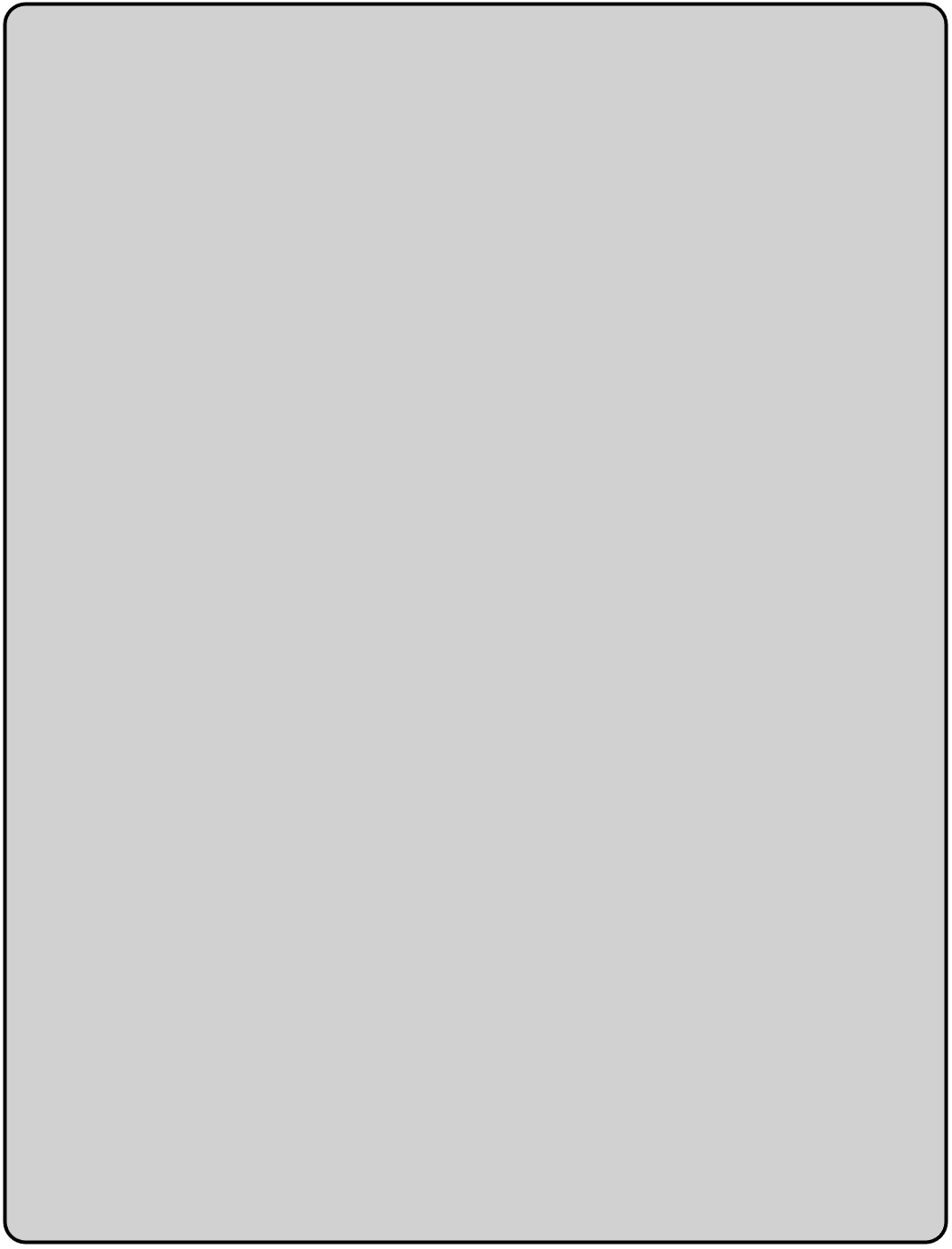}
        \put(3,49){\textcolor{black}{
        \begin{minipage}{0.455\linewidth}
            \fontsize{8}{12}\selectfont
            \textbf{\#\# Instructions \#\#}
            \\1. Repeat the question and then extract the objects mentioned in the question. Label the first object that appears as "Object A" and the second as "Object B".
            \\2. Describe the relative position \textcolor{red}{between Object B and A and between Object A and B.}
            \\3. Reference to the relationship between Object B and A and between Object A and B, and then \textcolor{blue}{think step by step} to use "yes" or "no" to answer the question.
            \\\textbf{\#\# Please output in the following format \#\#}
            \\Question:
            \\Object A:
            \\Object B:
            \\\textcolor{blue}{Horizontal relation} between Object B and A: B is <relation> A
            \\\textcolor{blue}{Vertical relation} between Object B and A: B is <relation> A
            \\\textcolor{blue}{Depth relation} between Object B and A: B is <relation> A
            \\\textcolor{blue}{Horizontal relation} between Object A and B: A is <relation> B
            \\\textcolor{blue}{Vertical relation} between Object A and B: A is <relation> B
            \\\textcolor{blue}{Depth relation} between Object A and B: A is <relation> B
            \\Thinking process: 
            \\Answer: 
            \\\textbf{\#\# Question \#\#}
            \\\{question\}
        \end{minipage}}}
    \end{overpic}
    \caption{Template prompt utilizing the bidirectional constraint. Prompting techniques are highlighted in blue. Terms related to the \textit{BA + AB} order are marked in red.} 
    \label{fig:prompt1}
\end{figure*}

\begin{figure*}[h]
    \centering
    \begin{overpic}[width=0.5\linewidth]{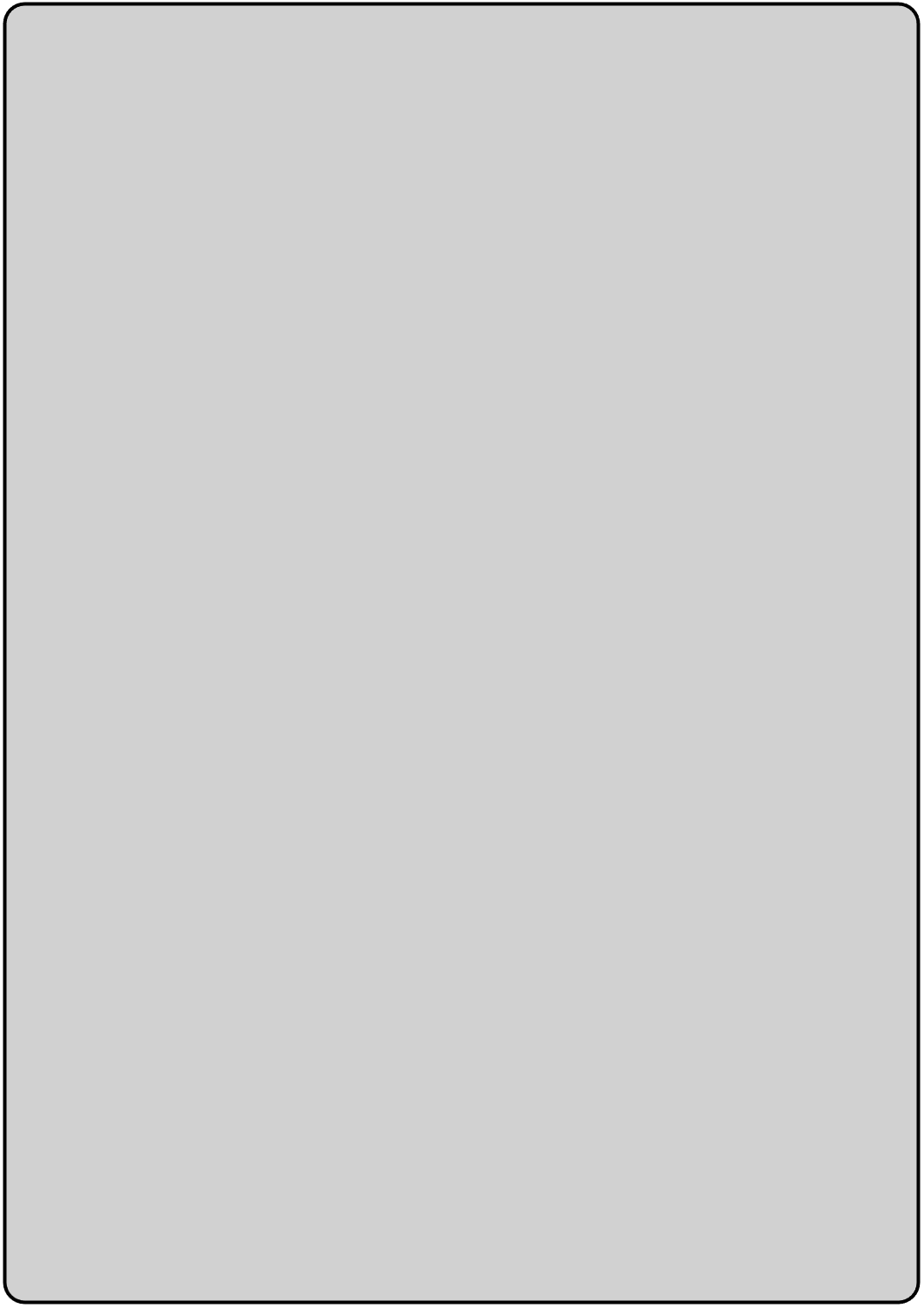}
        \put(3,49){\textcolor{black}{
        \begin{minipage}{0.455\linewidth}
            \fontsize{8}{12}\selectfont
            \textbf{\#\# Instructions \#\#}
            \\1. Repeat the question and then extract the objects mentioned in the question. Label the first object that appears as "Object A" and the second as "Object B". Select an object different from Object A or Object B in the image as "Object C".
            \\2. Describe the relative position \textcolor{red}{between Object A and C and between Object B and C.}
            \\3. Reference to the relationship between Object A and C and between Object B and C, and then \textcolor{blue}{think step by step} to use "yes" or "no" to answer the question.
            \\\textbf{\#\# Please output in the following format \#\#}
            \\Question:
            \\Object A:
            \\Object B:
            \\Object C:
            \\\textcolor{blue}{Horizontal relation} between Object A and C: A is <relation> C
            \\\textcolor{blue}{Vertical relation} between Object A and C: A is <relation> C
            \\\textcolor{blue}{Depth relation} between Object A and C: A is <relation> C
            \\\textcolor{blue}{Horizontal relation} between Object B and C: B is <relation> C
            \\\textcolor{blue}{Vertical relation} between Object B and C: B is <relation> C
            \\\textcolor{blue}{Depth relation} between Object B and C: B is <relation> C
            \\Thinking process: 
            \\Answer: 
            \\\textbf{\#\# Question \#\#}
            \\\{question\}
        \end{minipage}}}
    \end{overpic}
    \caption{Template prompt utilizing the transitivity constraint. Prompting techniques are highlighted in blue, and terms relevant to the reference relations (\textit{AC + BC} order) are highlighted in red.} 
    \label{fig:prompt2}
\end{figure*}

\begin{figure*}[h]
    \centering
    \begin{overpic}[width=0.5\linewidth]{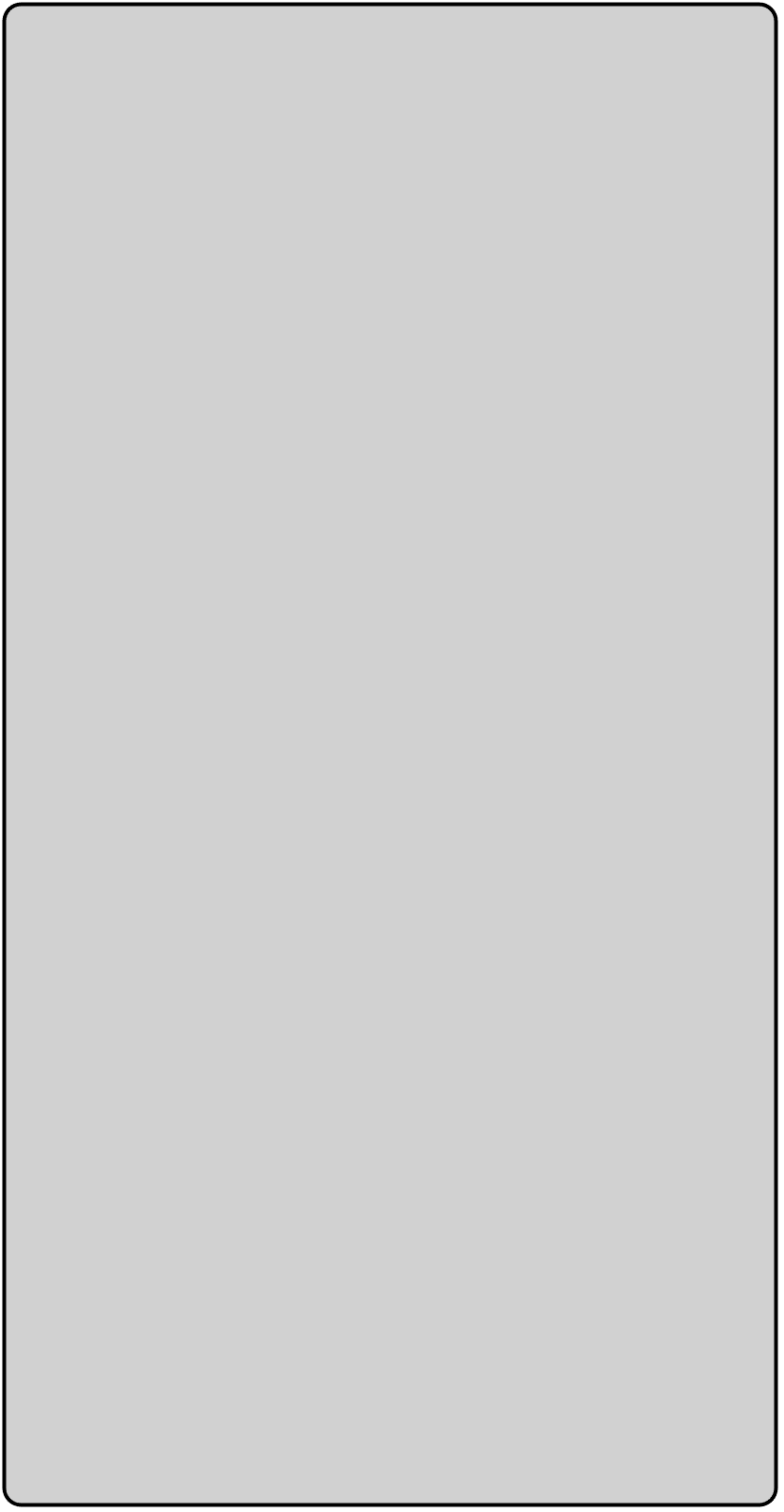}
        \put(2,49){\textcolor{black}{
        \begin{minipage}{0.455\linewidth}
            \fontsize{8}{12}\selectfont
            \textbf{\#\# Instructions \#\#}
            \\1. Repeat the question and then extract the objects mentioned in the question. Label the first object that appears as "Object A" and the second as "Object B". Select an object different from Object A or Object B in the image as "Object C"
            \\2. Describe the relative position \textcolor{red}{between Object A and C and between Object B and C.}
            \\3. Reference to the result of step 2 and image, describe the relative position \textcolor{orange}{between Object B and A and between Object A and B.}
            \\4. Reference to the relationship between Object B and A and between Object A and B, and then \textcolor{blue}{think step by step} to use "yes" or "no" to answer the question.
            \\\textbf{\#\# Please output in the following format \#\#}
            \\Question:
            \\Object A:
            \\Object B:
            \\Object C:
            \\\textcolor{blue}{Horizontal relation} between Object A and C: A is <relation> C
            \\\textcolor{blue}{Vertical relation} between Object A and C: A is <relation> C
            \\\textcolor{blue}{Depth relation} between Object A and C: A is <relation> C
            \\\textcolor{blue}{Horizontal relation} between Object B and C: B is <relation> C
            \\\textcolor{blue}{Vertical relation} between Object B and C: B is <relation> C
            \\\textcolor{blue}{Depth relation} between Object B and C: B is <relation> C
            \\\textcolor{blue}{Horizontal relation} between Object B and A: B is <relation> A
            \\\textcolor{blue}{Vertical relation} between Object B and A: B is <relation> A
            \\\textcolor{blue}{Depth relation} between Object B and A: B is <relation> A
            \\\textcolor{blue}{Horizontal relation} between Object A and B: A is <relation> B
            \\\textcolor{blue}{Vertical relation} between Object A and B: A is <relation> B
            \\\textcolor{blue}{Depth relation} between Object A and B: A is <relation> B
            \\Thinking process: 
            \\Answer: 
            \\\textbf{\#\# Question \#\#}
            \\\{question\}
        \end{minipage}}}
    \end{overpic}
    \caption{Template prompt utilizing the combined constraint. Prompting techniques are highlighted in blue, terms relevant to the transitivity constraint are highlighted in red, and terms relevant to the bidirectional constraint are highlighted in orange.} 
    \label{fig:prompt3}
\end{figure*}

\begin{figure*}[h]
    \centering
    \begin{overpic}[width=0.5\linewidth]{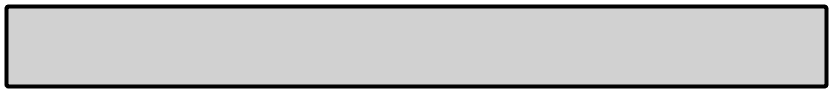}
        \put(4,4){\textcolor{black}{
        \begin{minipage}{0.455\linewidth}
            \fontsize{8}{12}\selectfont
            Use 'yes' or 'no' to answer the question: \{question\}
        \end{minipage}}}
    \end{overpic}
    \caption{Template prompt of Vanilla Baseline.} 
    \label{fig:prompt4}
\end{figure*}

\begin{figure*}[t]
    \centering
    \begin{overpic}[width=0.5\linewidth]{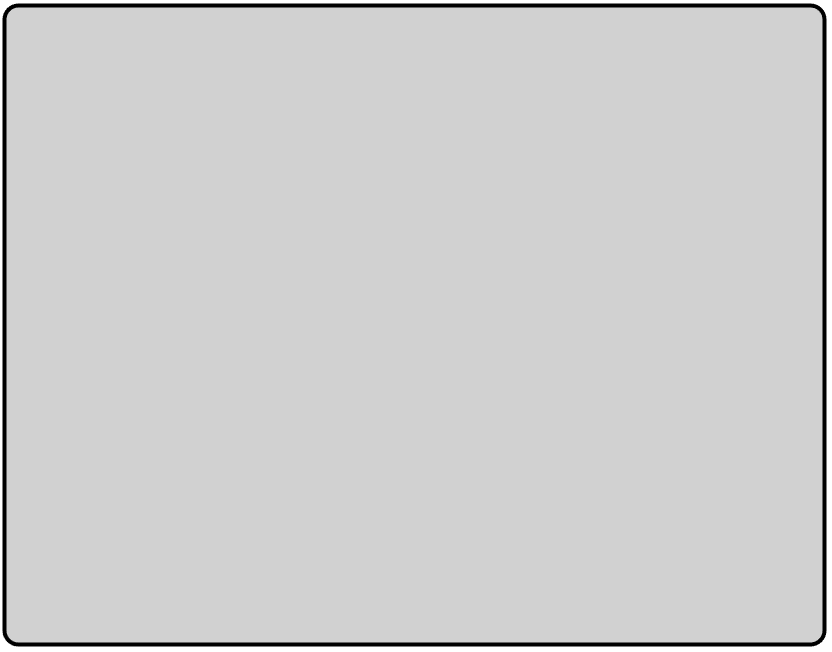}
        \put(4,38){\textcolor{black}{
        \begin{minipage}{0.455\linewidth}
            \fontsize{8}{12}\selectfont
            \textbf{\#\# Instructions \#\#}
            \\1. Repeat the question and then extract the objects mentioned in the question. Label the first object that appears as "Object A" and the second as "Object B".
            \\2. Think step by step to use 'yes' or 'no' to answer the question.
            \\\textbf{\#\# Please output in the following format \#\#}
            \\Question:
            \\Object A:
            \\Object B:
            \\Thinking process: 
            \\Answer: 
            \\\textbf{\#\# Question \#\#}
            \\\{question\}
        \end{minipage}}}
    \end{overpic}
    \caption{Template prompt of CoT+Structure Baseline.} 
    \label{fig:prompt5}
\end{figure*}

\begin{figure*}[h]
    \centering
    \begin{overpic}[width=0.5\linewidth]{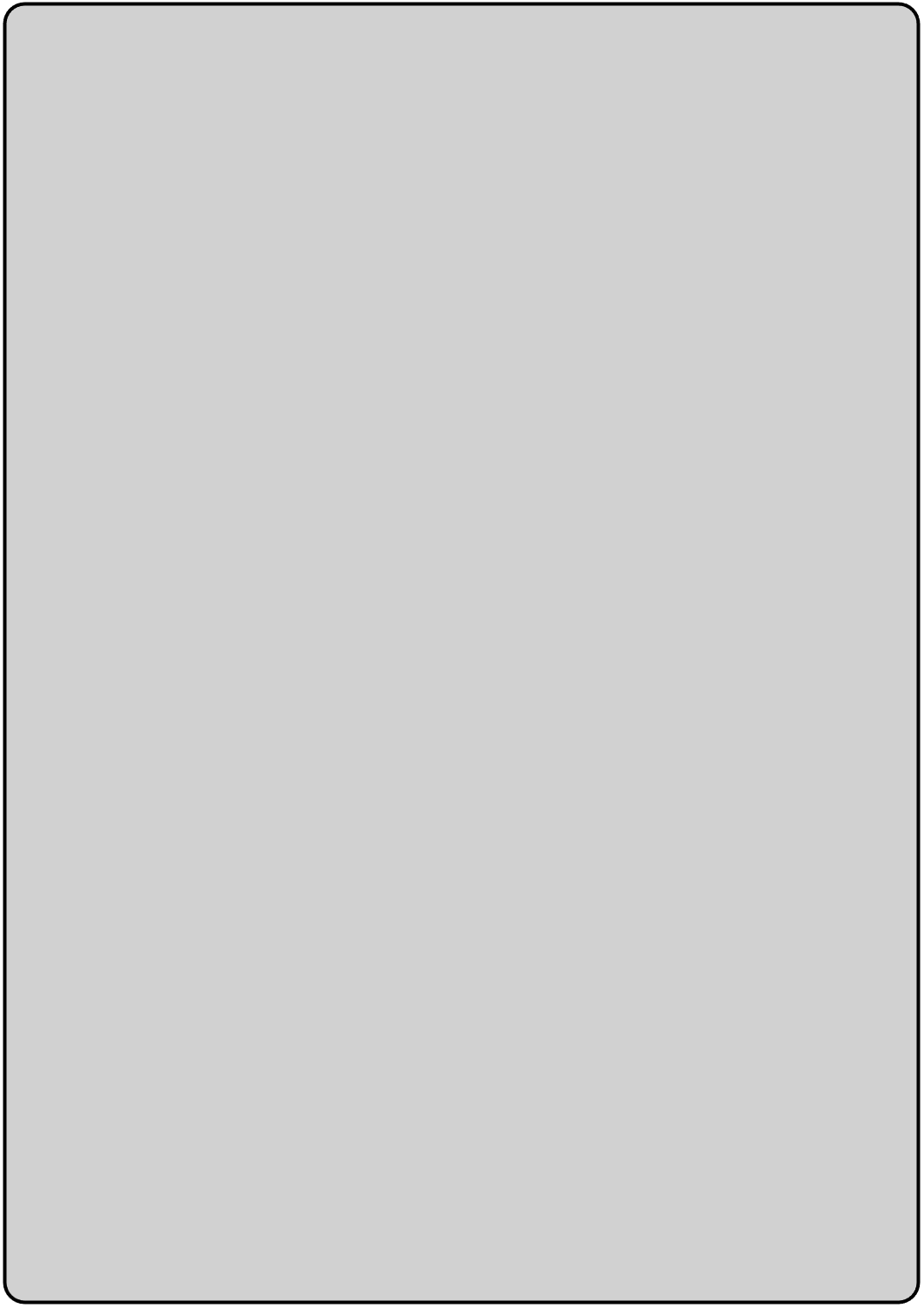}
        \put(3,49){\textcolor{black}{
        \begin{minipage}{0.455\linewidth}
            \fontsize{8}{12}\selectfont
            \textbf{\#\# Instructions \#\#}
            \\1. Repeat the question and then extract the objects mentioned in the question. Label the first object that appears as "Object A" and the second as "Object B". Select \textcolor{blue}{\{attribute\}} object different from Object A or Object B in the image as "Object C"
            \\2. Describe the relative position between Object A and C and between Object B and C.
            \\3. Reference to the relationship between Object A and C and between Object B and C, and then think step by step to use 'yes' or 'no' to answer the question.
            \\\textbf{\#\# Please output in the following format \#\#}
            \\Question:
            \\Object A:
            \\Object B:
            \\Object C:
            \\Horizontal relation between Object A and C: A is <relation> C
            \\Vertical relation between Object A and C: A is <relation> C
            \\Depth relation between Object A and C: A is <relation> C
            \\Horizontal relation between Object B and C: B is <relation> C
            \\Vertical relation between Object B and C: B is <relation> C
            \\Depth relation between Object B and C: B is <relation> C
            \\Thinking process: 
            \\Answer: 
            \\\textbf{\#\# Question \#\#}
            \\\{question\}
        \end{minipage}}}
    \end{overpic}
    \caption{Template prompt used in the reference  selection analysis of transitivity constraints. We replace \{attribute\} with the candidate attributes, such as "the largest" and "the most top."} 
    \label{fig:prompt6}
\end{figure*}

\begin{figure*}[h]
    \centering
    \begin{overpic}[width=0.5\linewidth]{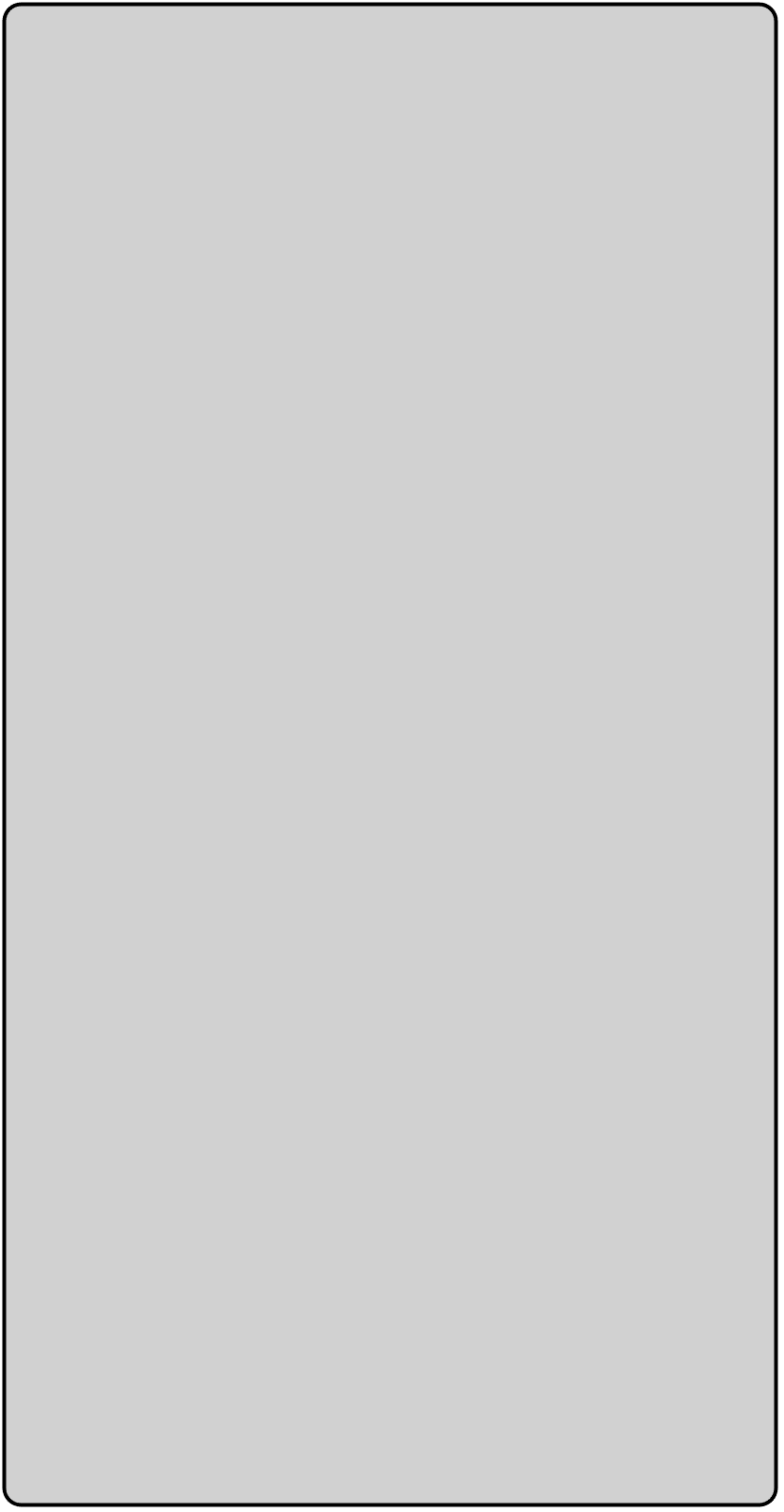}
        \put(2,49){\textcolor{black}{
        \begin{minipage}{0.455\linewidth}
            \fontsize{8}{12}\selectfont
            \textbf{\#\# Instructions \#\#}
            \\1. Repeat the question and then extract the objects mentioned in the question. Label the first object that appears as "Object A" and the second as "Object B". Select \textcolor{blue}{\{attribute\}} object different from Object A or Object B in the image as "Object C"
            \\2. Describe the relative position between Object A and C and between Object B and C.
            \\3. Reference to the result of step 2 and image, describe the relative position between Object B and A and between Object A and B.
            \\4. Reference to the relationship between Object B and A and between Object A and B, and then think step by step to use 'yes' or 'no' to answer the question.
            \\\textbf{\#\# Please output in the following format \#\#}
            \\Question:
            \\Object A:
            \\Object B:
            \\Object C:
            \\Horizontal relation between Object A and C: A is <relation> C
            \\Vertical relation between Object A and C: A is <relation> C
            \\Depth relation between Object A and C: A is <relation> C
            \\Horizontal relation between Object B and C: B is <relation> C
            \\Vertical relation between Object B and C: B is <relation> C
            \\Depth relation between Object B and C: B is <relation> C
            \\Horizontal relation between Object B and A: B is <relation> A
            \\Vertical relation between Object B and A: B is <relation> A
            \\Depth relation between Object B and A: B is <relation> A
            \\Horizontal relation between Object A and B: A is <relation> B
            \\Vertical relation between Object A and B: A is <relation> B
            \\Depth relation between Object A and B: A is <relation> B
            \\Thinking process: 
            \\Answer: 
            \\\textbf{\#\# Question \#\#}
            \\\{question\}
        \end{minipage}}}
    \end{overpic}
    \caption{Template prompt used in the reference  selection analysis of combined constraints.} 
    \label{fig:prompt7}
\end{figure*}

The template prompts utilizing bidirectional constraints, transitivity constraints, and combined constraints can be found in Figures~\ref{fig:prompt1},~\ref{fig:prompt2}, and~\ref{fig:prompt3} respectively. The template prompt of vanilla baseline is in Figure~\ref{fig:prompt4} and that of CoT+structure baseline is in Figure~\ref{fig:prompt5}. The template prompts used in the analysis of different reference selection strategies are in Figures~\ref{fig:prompt6} and~\ref{fig:prompt7}.

\end{document}